\pdfoutput=1

\documentclass[11pt]{article}

\usepackage[]{acl}

\usepackage{times}
\usepackage{latexsym}
\usepackage{soul}

\usepackage[T1]{fontenc}
\usepackage[utf8]{inputenc}

\usepackage{microtype}

\usepackage{inconsolata}

\usepackage{amsmath}
\usepackage{amssymb}
\usepackage{booktabs}
\usepackage{multirow}
\usepackage{graphicx}
\usepackage{caption}
\usepackage{subcaption}
\usepackage{color, colortbl}
\usepackage{mathtools}
\usepackage{xspace}
\usepackage{arydshln}  %

\usepackage{todonotes}
\makeatletter
\newcommand*\iftodonotes{\if@todonotes@disabled\expandafter\@secondoftwo\else\expandafter\@firstoftwo\fi}  %
\makeatother

\usepackage{colortbl}%
\usepackage[normalem]{ulem}

\newcommand{\mtdata}{{\sf MAPLE}\xspace}

\usepackage{CJKutf8}
\newcommand{\chinese}[1]{{\begin{CJK*}{UTF8}{gkai} #1 \end{CJK*}}}

\definecolor{bblue}{HTML}{4F81BD}
\definecolor{rred}{HTML}{c4260b}
\definecolor{ggreen}{HTML}{098c1f}
\DeclareRobustCommand{\hlred}[1]{{\textcolor{rred}{#1}}}
\DeclareRobustCommand{\hlblue}[1]{{\textcolor{bblue}{#1}}}
\DeclareRobustCommand{\hlgreen}[1]{{\textcolor{ggreen}{#1}}}
\DeclareMathOperator{\sech}{sech}

\title{A Preference-driven Paradigm for Enhanced Translation with Large Language Models}

\author{Dawei Zhu$^{1,2}$\thanks{\hspace{1.5 mm}Work done during an internship at Amazon.} \, Sony Trenous$^{1}$ \, Xiaoyu Shen$^{1}$\thanks{\hspace{1.5 mm}Now with Eastern Institute of Technology, Ningbo.} \, Dietrich Klakow$^{2}$ \, \\\textbf{Bill Byrne}$^{1}$ \, \textbf{Eva Hasler}$^{1}$ \\
$^1$Amazon AGI\\
$^2$Saarland University, Saarland Informatics Campus\\
\texttt{\{daweizhu,trenous,willbyrn,ehasler\}@amazon.com}}

\begin{document}
\maketitle
\begin{abstract}
Recent research has shown that large language models (LLMs) can achieve remarkable translation performance through supervised fine-tuning (SFT) using only a small amount of parallel data. However, SFT simply instructs the model to imitate the reference translations at the token level, making it vulnerable to the noise present in the references. Hence, the assistance from SFT often reaches a plateau once the LLMs have achieved a certain level of translation capability, and further increasing the size of parallel data does not provide additional benefits. To overcome this plateau associated with imitation-based SFT, we propose a preference-based approach built upon the Plackett-Luce model. The objective is to steer LLMs towards a more nuanced understanding of translation preferences from a holistic view, while also being more resilient in the absence of gold translations. We further build a dataset named \mtdata to verify the effectiveness of our approach, which includes multiple translations of varying quality for each source sentence. Extensive experiments demonstrate the superiority of our approach in ``breaking the plateau'' across diverse LLMs and test settings. Our in-depth analysis underscores the pivotal role of diverse translations and accurate preference scores in the success of our approach.\footnote{The \mtdata dataset is available at: \url{https://github.com/amazon-science/preference-driven-mt}}
\end{abstract}

\section{Introduction}
The emergence of Large Language Models (LLMs) has significantly transformed the landscape of NLP, showcasing outstanding capabilities in a spectrum of NLP tasks~\cite{Brown2020_gpt3, Scao2022_bloom, Chowdhery2023_palm, Touvron2023_llama}. This transformation extends to machine translation (MT)~\cite{OpenAI2023_gpt4,Jiao2023_gpt4_rivals_google_deepl, Hendy2023_chatgpt_translation}. Through supervised fine-tuning (SFT) using a small amount of parallel data, LLMs demonstrate the capability to compete with established commercial translation services such as Google Translate, particularly in high-resource languages~\cite{Jiao2023_parrot, Zhang2023_mt_qlora}.

Nevertheless, SFT trains the model to imitate reference translations token by token, making it vulnerable to the noise present within the data~\cite{Ott2018, Zhou2023_lima, Touvron2023_llama2}. The noise can stem not only from the lack of attention by annotators, but also from the inherent challenge of achieving perfect translations due to the intricate interplay of language, culture, and vocabulary. As an adept translator requires not only linguistic proficiency but also a deep understanding of cultural contexts and nuances in both the source and target, it is nearly unattainable to gather extensive parallel translations of top-notch quality~\cite{Khayrallah2018_translation_noise,Herold2022,Maillard2023}.
As a result, the performance enhancement achieved through SFT often quickly reaches a plateau. Further increasing the volume of parallel translations typically yields minimal additional benefits, and may instead impair the translation capabilities of LLMs~\cite{Xu2023_alma}.

To alleviate aforementioned limitation of SFT, endeavors have been made to provide LLMs with holistic assessment of contrasting examples rather than token-level imitations. \citet{Jiao2023_parrot, Chen2023_SWIE} add a flawed translation to the reference translation in the model input, encouraging the target LLM to recognize their quality difference. \citet{Zeng2023_tim} also use a pair of translations, but they additionally optimize the LLM to favor better translations through ranking loss. Nevertheless, these works have shared limitations. First, the flawed translations are either generated by adding artificial noise to the reference translations or by other (smaller) MT systems. These imperfections in translations can be obvious and easy for LLM to distinguish, weakening the learning signal. Second, they only provide the relative ranking of the two translations, without quantifying the extent of their quality differences.

In this work, we present a framework based on the Plackett-Luce model to explicitly align the generation probability of the target LLM with human preferences~\cite{Plackett1975_pl_model}. Instead of using artificial noise, we collect contrasting translations generated by our target LLM, directing our optimization efforts toward ``hard negative examples''~\cite{Robinson2021}. Human preferences are denoted with precise scores rather than general ranking orders to teach LLMs about the nuances in different translations. LLMs are then trained to enhance their capabilities incrementally from the learnt nuances without depending solely on the existence of ``gold references'', so as to effectively break the plateau associated with SFT. 

We build a dataset, which we refer to as \mtdata, to facilitate preference learning. It equips each source sentence with five translations in diverse quality, scored by professional translators. By performing preference learning on \mtdata, our final MT model outperforms other MT models based on the same foundation LLM by up to 3.96 COMET score.  We further show that while the intention of creating \mtdata is to enhance our target LLM, it can be reused to improve other LLMs, helping them break the performance plateau with up to 1.4M parallel data. Finally, we analyze the key factors that make preference learning effective.

Our contributions are as follows. \textbf{(1)} We leverage preference learning to teach LLMs a holistic notion of translation quality. Extensive experiments show that our model consistently outperforms strong baselines on two test sets across four translation directions. \textbf{(2)} We revisit the underlying modelling assumptions leading to the Bradley-Terry and Plackett-Luce ranking models and discuss how preference distances can be incorporated directly into the ranking models. \textbf{(3)} We meticulously construct an MT-oriented preference dataset, \mtdata, employing professional human translators to obtain quality scores for multiple translations corresponding to the same source sentence. We release our dataset to facilitate future MT research. \textbf{(4)} Our in-depth analysis reveals that high-contrast pairs and accurate quality scores are crucial in enhancing the effectiveness of our approach, providing guidance for maximizing the benefits of preference learning.

\section{Related Work}
\paragraph{LLM-based MT.} One simple and effective approach to use LLMs for translation tasks is through prompting. Research in this field involves examining the impact of model sizes, the number of examples (``shots'') used, and template choices~\cite{Zhang2023,Bawden2023_bloom_mt,Mu2023,zhang2024impact}. Moreover, \cite{Ghazvininejad2023, He2023} highlight that better translations can be achieved by adding supplementary information to prompts, or engaging LLMs in related tasks prior to translation. Alternatively, another research direction seeks to fully tailor LLMs for MT tasks. \citet{Jiao2023_parrot, Zeng2023_tim, Chen2023_SWIE, Alves2023, Zhang2023_mt_qlora} further train LLMs on parallel data via (parameter-efficient) fine-tuning. \citet{Xu2023_alma} show that increasing the size of parallel data may not further improve LLM. The diminished returns from increasing data volume are likely due to data noise. Recent analyses suggest that quality trumps quantity when it comes to data effectiveness~\cite{Zhu2023_weaker_than_you_think, Zhou2023_lima}. Leveraging these insights, we goes beyond merely fitting the reference translations. Instead, we aim to enhance the LLM's ability to discern translations of varying quality, encouraging the generation of more precise translations while suppressing flawed outputs.

\paragraph{Human preference alignment.} \citet{Ouyang2022_instructGPT} align LLMs with human intentions and values by training a reward model for preference ranking and optimizing the LLMs through the PPO algorithm~\cite{Schulman2017_ppo}. However, the online reinforcement learning nature of PPO leads to considerable computational costs and is known for its sensitivity to hyperparameters~\cite{Islam2017, Huang2022}. To ease the alignment, \citet{Hu2023_align_llm_offline, Dong2023_SteerLM} suggest offline RL algorithms where samples are pre-generated. Further research goes a step beyond by directly employing the target LLMs as reward models. \citet{Yuan2023_rrhf} use a ranking loss to steer LLMs towards generating helpful responses and avoiding harmful ones. In a similar vein, \citet{Rafailov2023_dpo, Song2023_pro, Hejna2023_cpl} use the Plackett-Luce model (Plackett, 1975) to capture human preferences in alignment. In this work, we adopt the Plackett-Luce model to MT, teaching the model to discern nuances in different translations and to prefer accurate translations.

\section{Methodology}
We aim to enhance LLM in MT tasks via a two-stage optimization process. We first fine-tune the target LLM with a small set of high-quality parallel data to elicit its translation ability (Section~\ref{sec:sft}). This mirrors the supervised fine-tuning approach used in prior work, where LLMs were tailored to follow instructions~\cite{Taori2023_alpaca, Zheng2023_vicuna}. We then use preference learning to guide the LLM to prioritize the generation of accurate translations over flawed ones (Section~\ref{sec:preference_learning}).

\subsection{Supervised fine-tuning}
\label{sec:sft}
We begin with optimizing our target LLM on parallel data to specialize it for  translation. Let $x$ and $y$ denote the source and target sentence, respectively. Following \citet{Jiao2023_parrot} we first construct a prompt by applying an instruction template $\mathcal{I}$ to $x$. The instruction template is randomly sampled from an instruction pool for each training sample. The target LLM, denoted by $\pi_{\theta}$ is optimized through the log-likelihood loss:
\begin{align}
    \label{eq:sft}
    \mathcal{L}_{SFT}(\pi_\theta) & = -\log \pi_{\theta}(x, y) \nonumber \\
    & = -\sum_{t} \log P_{\pi_{\theta}}(y_{t}|y_{1, \cdots, t-1}, \mathcal{I}(x)) 
\end{align}
where $\pi_{\theta}(x, y)$ denotes the likelihood of $\pi_{\theta}$ generating output $y$ given input $x$. Note that in a standard implementation, a decoder-only LLM will also predict tokens within $\mathcal{I}(x)$, we zero-out the loss on these tokens as our main goal is to teach translation, not to model the input distribution~\cite{Touvron2023_llama2}.\footnote{
As per \citet{Ouyang2022_instructGPT}, we use the term ``SFT'' which is interchangeably referred to as ``instruction-tuning'' or simply ``fine-tuning'' in current literature to convey the same concept.}

\subsection{Preference learning}
\label{sec:preference_learning}
The goal of the preference learning stage is to explicitly optimize the target LLM to favor accurate translations over erroneous ones. Formally, consider a set of translations ${y^1, \cdots, y^L}$ corresponding to a source sentence $x$. We assume that these translations are ordered by preference: $y^i \succ_{x} y^j$ for $i<j$. That is, translation $y^i$ is preferred over $y^j$ as a translation of the source sentence $x$. We further assume that there is some underlying reward model $r^{*}$ that reflects the quality of the translations, which we cannot access but which we can approximate. Under the Plackett-Luce ranking model \cite{Plackett1975_pl_model}, the distribution of preferences can be formulated as follows:
\begin{equation}
\begin{aligned}
\label{eq:plackett-luce}
p^{*}(y^{1:L}_{\succ_{x}}| x) = \prod_{i=1}^{L-1}\frac{\exp(r^{*}(x, y^{i}))}{\sum_{j=i}^{L} \exp(r^{*}(x, y^{j}))}
\end{aligned}
\end{equation}
where $y^{1:L}_{\succ_{x}}$ is a shorthand for the complete preference ranking $y^1 \succ_{x}, \cdots, \succ_{x} y^L$. In practice, given a training set $\mathcal{D}$ with translations equipped with a preference ranking, a reward model $r_{\theta}$ can be trained via maximum likelihood estimation~\cite{Cheng2010}:
\begin{align}
\label{eq:plackett-luce-loss}
\mathcal{L}_{PL}(r_\theta) = - &\mathbb{E}_{x, y^{1:L}_{\succ_{x}}\in \mathcal{D}}\sum_{i=1}^{L-1} \left[ r_{\theta} (x, y^{i}) - \right. \nonumber \\
&  \big. \log\sum^{L}_{j=i} \exp(r_{\theta}(x, y^{j})) \big]
\end{align}
Following recent work \cite{Rafailov2023_dpo, Song2023_pro, Hejna2023_cpl}, we parameterize the reward model  using the target LLM $\pi_{\theta}$ and rewrite the above objective as:
\begin{equation}
\label{eq:plackett-luce-rewrite}
\mathcal{L}_{PL}(\pi_\theta) = -\mathbb{E}_{x, y^{1:L}_{\succ_{x}} \in \mathcal{D}}\sum_{i=1}^{L-1} \log \frac{\pi_{\theta}(x, y^i)}{\sum_{j=i}^{L}\pi_{\theta}(x, y^j)}
\end{equation}
where $r_{\theta} \coloneqq \log(\pi_{\theta})$. Through optimizing Equation~\ref{eq:plackett-luce-rewrite}, we explicitly align the LLM generation probability with the translation quality. 

A caveat when optimizing Equation~\ref{eq:plackett-luce-rewrite} is that the ranking information omits any measure of absolute translation quality, which may lead to inadvertent suppression of the likelihood of good translations. Consider a case where we have a pair of translations, $y^1$ and $y^2$, which are both acceptable translations but have different word orders that causes minor difference in preference. Optimizing Equation~\ref{eq:plackett-luce-rewrite} may cause the model to raise the probability of  $y^1$ and to suppress the probability $y^2$, which may damage the model.\footnote{\citet{Cheng2008} show that, while the preference can be learned asymptotically solely through ranking information, incorporating additional, more detailed, preference information (e.g., distance) makes the learning process more data-efficient. Table~\ref{tab:pl-ablation-study} presents an ablation study.} To address this issue, we follow \citet{Song2023_pro} to consider the preference distance in $\mathcal{L}_{PL}$:
\begin{align}
    \label{eq:plackett-luce-loss-with-distance}
    \mathcal{L}_{PLD}(\pi_\theta) & = \nonumber \\
    & - \mathbb{E}_{x, y^{1:L}_{\succ_{x}} \in \mathcal{D}}\sum_{i=1}^{L-1} \log \frac{\pi^{d^{i}_{i}}_{\theta}(x, y^i)}{\sum_{j=i}^{L}\pi^{d^{j}_{i}}_{\theta}(x, y^j)} \nonumber \\
    \text{where} & \nonumber  \\
    d^{j}_{i} &= r^{*}(x, y^{i}) - r^{*}(x, y^{j}), \text{for } j>i \nonumber \\
    d^{i}_{i} &= \max_{j>i} (d^{j}_{i})
\end{align}
We obtain the ground truth preference value $r^{*}(x, y)$ through human annotation, which will be detailed in Section~\ref{sec:preference_collection}. Finally, we combine a SFT loss calculated on the best translation $y^{1}$ with $\mathcal{L}_{PLD}$, making the complete loss function:
\begin{equation}
\label{eq:plackett-luce-combined-loss}
\mathcal{L} = \mathcal{L}_{PLD} + \beta \mathcal{L}_{SFT}
\end{equation}
where the hyperparameter $\beta$ balances the strengths of preference learning and SFT. We use \textbf{PL} as an abbreviation of our preference learning method (i.e., optimizing Equation~\ref{eq:plackett-luce-combined-loss}) in the subsequent text.

We now provide some justification for directly incorporating  preference distances into the Plackett-Luce model by studying the original derivation of the binary case ($L=2$)~\cite{Thurstone1927, Mosteller1951, Bradley1953,hamilton_many_2023}.     Denote the preferences  for $y^{i}$ and $y^{j}$ by random variables $X_i$ and $X_j$ such that the probability that $y^i$ is preferred to $y^j$ is  $\pi_{ij} = P(X_i > X_j)$. Assuming that $X_i$ and $X_j$ follow Gumbel distributions\footnote{Assuming preferences arise from a large number of i.i.d. contributions, a normal distribution results in the limit if these are averaged while the Gumbel distribution results from taking their maximum~\cite{hamilton_many_2023}.} with locations $s_i$ and $s_j$ and a common scale parameter $\gamma$, the difference between the two random variables $d_{ij} = X_i - X_j$ follows a logistic distribution with location $s_i - s_j$ and scale $\gamma$:
\begin{align}
    \label{eq:preference-distance-distribution}
    d_{ij} \sim  \frac{1}{4\gamma} \sech ^2 (\frac{d_{ij} - (s_i - s_j)}{2 \gamma})
\end{align}
By defining $\pi_{i}=e^{s_i}$, it follows that 
\begin{align}
    \label{eq:preference-positive-distance}
    \pi_{ij} &= P(d_{ij} > 0) \nonumber\\
    &= \int_0^\infty \frac{1}{4\gamma} \sech^2 (  \frac{d_{ij} - (s_i - s_j)}{ 2 \gamma}  ) d d_{ij} \nonumber \\
    &= \frac{\pi_i^{\frac{1}{\gamma}}}{\pi_i^{\frac{1}{\gamma}} + \pi_j^{\frac{1}{\gamma}}}
\end{align}
Usually  the scale parameter $\gamma$  is set to $1$ which yields the Bradley-Terry model \cite{Bradley1952} (and Equation 13 of \citet{Bradley1953}).

To introduce distance information for the binary preference case, we first note that $d_1^1 = d_1^2$ for $L=2$ 
(from Equation~\ref{eq:plackett-luce-loss-with-distance}).    We then  take $\gamma=\frac{1}{d_1^2}$ and $\pi_i = \pi_{\theta}(x, y^{i})$, which yields:
\begin{equation}
    \label{eq:preference-distance-binary-case}
\pi_{12} = \frac{\pi_\theta^{d_1^2}(x,y^1)}{\pi_\theta^{d_1^1}(x,y^1) + \pi_\theta^{d_1^2}(x,y^2)} 
\end{equation}
This shows that, for the binary case,  preference distances based on the ground truth preferences can be incorporated exactly into the Bradley-Terry distribution  by assuming that the $X_1$ and $X_2$ have Gumbel distributions with location parameters $s_i = \log \pi_\theta(x, y^i)$ and scale parameter $\gamma = \frac{1}{r^\ast(x,y^1) - r^\ast(x,y^2)}$.

We derive and discuss the more general case of Equation~\ref{eq:plackett-luce-loss-with-distance} ($L>2$) in Appendix~\ref{sec:appx:incoerpate_with_more_than_2_preferences}.

\paragraph{Connections with DPO} The preference learning framework investigated here shares a common origin with DPO~\cite{Rafailov2023_dpo} in the Bradley-Terry and Plackett-Luce models over rankings (Equation~\ref{eq:plackett-luce}, and Equation 18 of \citet{Rafailov2023_dpo}).  Here, the target LLM $\pi_\theta$ serves directly as the reward function ($r_\theta = \log(\pi_\theta))$, whereas the DPO reward function also includes a reference distribution $\pi_{ref}$ that arises from the KL-divergence constraint term in its RL objective function. By contrast, regularization in this work is through an external SFT term (Equation~\ref{eq:plackett-luce-combined-loss}) distinct from the reward function. We note also that the use of distance functions based on ground truth reference values brings additional information into our ranking model beyond preference order alone.

\section{Human preference data collection}
\label{sec:preference_collection}
We build \mtdata (\textbf{MA}chine translation dataset for \textbf{P}reference \textbf{LE}arning), a dataset derived from WMT20/21 test sets. It contains multiple translations per source sentence, each assigned a real-valued human preference score. \mtdata covers four translation directions: German-to-English (\texttt{de}$\rightarrow$\texttt{en}), Chinese-to-English (\texttt{zh}$\rightarrow$\texttt{en}), English-to-German (\texttt{en}$\rightarrow$\texttt{de}), and English-to-Chinese (\texttt{en}$\rightarrow$\texttt{zh}). For each direction, 1.1K source sentences are sampled from the test sets of WMT20/21. Each source sentence is associated with five translations, including one reference translation from WMT20/21, and four translations generated by VicunaMT, our target LLM that we aim to improve through preference learning (see training details of VicunaMT in Section~\ref{sec:sft_makes_good_mt_models}). Among the four translations, one is generated using beam search with a beam size of four, and three translations are obtained through nucleus sampling~\cite{Holtzman2020} with $p=0.9$. We also build a development set containing 200 source sentences per direction sourced from News Crawl 2022. Altogether, \mtdata contains 5.2K source sentences and 26K translations with preference scores. See Appendix~\ref{sec:appx:maple_data_construction} for more detail on the translation collecting process.
\begin{figure}[h]
    \centering
     \includegraphics[width=\columnwidth]{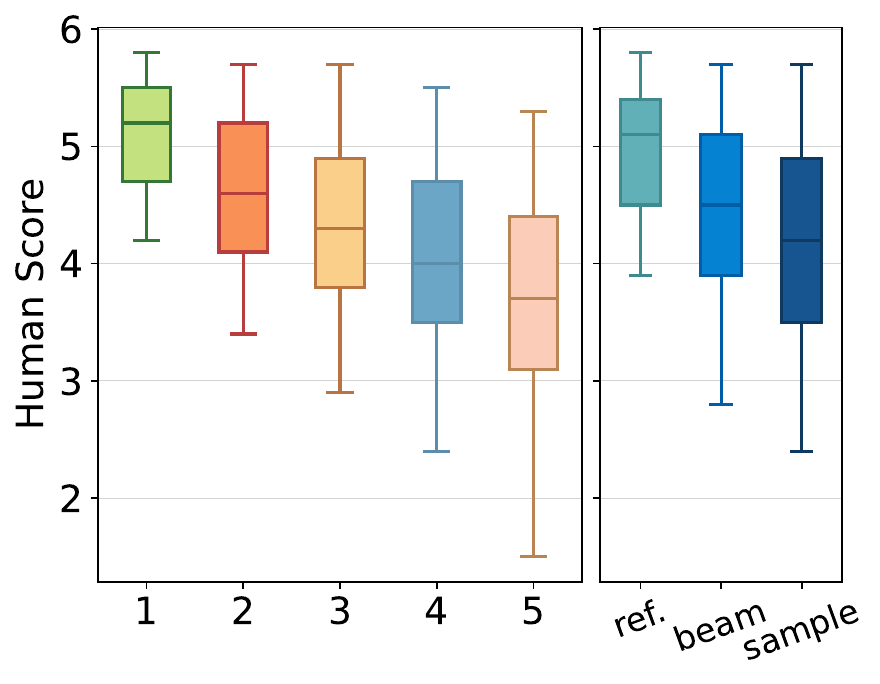}
    \caption{Human score distribution of translations by rank (left) and source (right).}
    \label{fig:dataset_statistics}
\end{figure}

\begin{table}[h]
    \centering
    \small
    \begin{tabular}{p{1.6cm}p{4.9cm}}
        \toprule
        Source & Zu einem großen Tuning-Treffen ist es am Samstagabend (25. Juli 2020) in Nürnberger \hlblue{Südstadt} gekommen. \\
         & \textit{(A large tuning meeting took place on Saturday evening (July 25, 2020) in Nuremberg's Südstadt district.) }\\
        \midrule
        Reference translation & A large tuning meetup took place in \hlred{a city south of} Nürnberg this Saturday evening. \\
        \midrule
        Best  & On Saturday evening (25th July 2020)\\
        translation  &  a large tuning meeting took place in Nuremberg's \hlgreen{south district}. \\
        \bottomrule
    \end{tabular}
    \caption{An example where the reference translation is less accurate than the best model prediction. More examples are in Appendix~\ref{sec:appx:more_details_on_our_data_examples}.}
    \label{tab:model-better-than-reference}
\end{table}

\paragraph{Annotation guidance.}  We send both the source sentence and the corresponding five translations to a panel of translators for evaluation. Each example (source sentence and its translations) is assigned to two different professional translators. They observe the source and the five translations at the same time, and assign scores between 1 (worst) and 6 (best) in increments of 0.2 using a slider. See Appendix~\ref{sec:appx:scoring_rubric} for the full scoring rubric.

\paragraph{Dataset statistics.} The score distribution is shown in Fig.~\ref{fig:dataset_statistics}. The left side shows the score distribution by rank, and we can see \mtdata contains translations that exhibit a wide range of qualities. The right side shows the score distribution by translation type, and as expected the reference is ranked highest, followed by the beam search and the nucleus samples. Nonetheless, there is considerable overlap in the score distributions, and we find that in 21\% of the cases, the beam search predictions are scored higher than the reference translation. 
Table~\ref{tab:model-better-than-reference} shows an example where the reference translation contains an error.

\section{Experiments}
In this section, we present our MT model trained using the proposed two-stage framework and compare it with strong LLM-based MT systems. 

\paragraph{Datasets.} We train and evaluate the model on data on four translation directions: \texttt{en}$\leftrightarrow$\texttt{de} and \texttt{en}$\leftrightarrow$\texttt{zh}.  In the SFT stage, we use high-quality test sets from WMT17/18/19 for training, containing 30K parallel sentences in total across the four directions. The WMT21 test set is used for validation. In the preference learning stage, we train on \mtdata, and validation is done on the remaining data from WMT20/21 test sets which was not selected for inclusion in \mtdata. We evaluate trained models on the test sets of WMT22~\cite{Kocmi2022_wmt22} and FLORES-200~\cite{Costajussa2022_nllb}. Refer to Appendix~\ref{sec:appx:dataset_statistics} for detailed data statistics.
\paragraph{Training.}In both SFT and PL stages, we use a learning rate of 5e-6, an effective batch size of 96, and a linear learning rate schedule with a warmup ratio of 0.1. For each training instance, one MT instruction is randomly selected from an instruction pool containing 31 MT instructions. See Appendix~\ref{sec:appx:instruction_pool} for the complete list of instructions.
\paragraph{Evaluation.} At inference time, a fixed MT translation instruction is used. The maximum generation length is set to 512. We use a beam size of 4 for decoding and report BLEU~\cite{Papineni2002_bleu} and COMET~\cite{Rei2022_comet} scores.

\subsection{SFT makes good translation models}
\label{sec:sft_makes_good_mt_models}
The SFT stage seeks to train a well-performing foundation MT model using parallel data. When applying SFT, we can either select a pre-trained LLM, or its instruction-tuned version. Prior research uses both types of LLMs interchangeably, leaving it unclear which is preferable in practice. To address this gap, we explore three popular families of open-access LLMs, performing SFT on both their raw (i.e., only pre-trained) and instructed-tuned versions. Specifically, we consider LLaMA-1 \cite{Touvron2023_llama}, Mistral \cite{Jiang2023_mistral} and BLOOM \cite{Scao2022_bloom}; and their instruction-tuned versions, which are Vicuna \cite{Zheng2023_vicuna}, Mistral-Instruct, and BLOOMZ \cite{Muennighoff2023_bloomz}. The 7B parameter variants of these models are used here.

\begin{table}[h]
\resizebox{\columnwidth}{!}{
\begin{tabular}{@{}lccccc@{}}
\toprule
                         & \multicolumn{1}{c}{\texttt{de}$\rightarrow$\texttt{en}} & \multicolumn{1}{c}{\texttt{en}$\rightarrow$\texttt{de}} & \texttt{en}$\rightarrow$\texttt{zh}                     & \multicolumn{1}{c}{\texttt{zh}$\rightarrow$\texttt{en}} & Avg.  \\ \midrule
                         & \multicolumn{5}{c}{\textit{WMT22}}                                                                                             \\
BLOOM                    & 49.86                     & \multicolumn{1}{c}{41.95} & 51.59                     & 55.21                     & 49.65 \\
\multicolumn{1}{@{}r}{\textit{+SFT}} & 77.21 & \multicolumn{1}{c}{69.17} & 84.60      & 78.76   & 77.44 \\
\rowcolor{blue!5}
BLOOMZ   & 74.58       & \multicolumn{1}{c}{62.52} & 83.10           & 78.29          & 74.62 \\
\rowcolor{blue!5}
\multicolumn{1}{@{}r}{\textit{+SFT}} & 77.24  & \multicolumn{1}{c}{69.32} & \textbf{84.95}  & \textbf{78.77} & 77.57 \\
Mistral &       54.18                    &         49.08                  &         49.10                  &           55.47                &   51.96    \\
\multicolumn{1}{@{}r}{\textit{+SFT}} & 83.15                     & 81.10                     & 81.48                     & 78.05                     & 80.95 \\
\rowcolor{blue!5}
Mistral-Ins.              & 82.45                     & 80.39                     & 76.57                     & 77.73                     & 79.28 \\
\rowcolor{blue!5}
\multicolumn{1}{@{}r}{\textit{+SFT}} & 82.68                     & 81.23                     & 82.49 & 77.73                     & 81.03 \\
LLaMA-1                    &       63.29                    &           55.29                & 45.80      &         55.17                  &   54.89    \\
\multicolumn{1}{@{}r}{\textit{+SFT}} & 83.30                     & 82.54                     & 77.58 & 75.78                     & 79.80 \\
\rowcolor{blue!5}
Vicuna                   &         82.55                  &         82.02                  & 81.42     &           74.81              &   80.20    \\
\rowcolor{blue!5}
\multicolumn{1}{@{}r}{\textit{+SFT}} & \textbf{83.55}                     & \textbf{82.79}                     & \multicolumn{1}{c}{81.27} & 77.39                     & \textbf{81.25} \\
\midrule
& \multicolumn{5}{c}{\textit{FLORES-200}}                                                                                             \\ 
BLOOM                    & 55.03                         & 42.36                         & 53.82                                                               & 60.25                         & 52.86                         \\
\multicolumn{1}{@{}r}{\textit{+SFT}} & 83.69                         & 67.43                         & \textbf{86.06}                                                               & \textbf{85.45}                         & 80.66                         \\
Mistral                  &          42.36                       &                32.74                 &                33.35                                                       &    42.10                             &          37.64                       \\
\multicolumn{1}{@{}r}{\textit{+SFT}} &            88.63                     &       84.49                          &                        80.97                                               &                  85.17              &               84.81                  \\
\rowcolor{blue!5}
Mistral-Ins.              &       88.04                          &             82.55                    &                      73.20                                                 &         83.70                        &          81.87                       \\
\rowcolor{blue!5}
\multicolumn{1}{@{}r}{\textit{+SFT}} &          88.21                       &        83.73                         &               82.41                                                        &                    84.77             &                84.78                 \\
LLaMA-1                    &            58.89                     &            52.71                     &             42.77                  &                49.92                        &         51.07                                                         \\
\multicolumn{1}{@{}r}{\textit{+SFT}} &             88.50                    &                    84.82             &                                        76.73                               &          83.09                       &              83.29                   \\
\rowcolor{blue!5}
Vicuna     &     87.82       &    84.17     &    81.52        &    81.53   &        83.76                         \\
\rowcolor{blue!5}
\multicolumn{1}{@{}r}{\textit{+SFT}} &     \textbf{88.66}                           &              \textbf{86.27}                    &                       80.62                                                &                    84.44             &               \textbf{85.00}                  \\
\bottomrule
\end{tabular}
}
\caption{Model performance (in COMET score) before and after performing SFT on parallel data. Rows in blue indicate instruction-tuned LLMs. Best results are in \textbf{bold}. Instruction-tuned LLMs yield high COMET scores even without SFT. Raw LLMs benefit the most from SFT. Vicuna performs the best on average on both test sets. We exclude BLOOMZ on FLORES-200 as it is a part of BLOOMZ's training data. Performance measured by BLEU score is reported in Appendix~\ref{sec:appx:more_results_in_sft_stage}. }
\label{tab:sft-performance-wmt22-comet}
\end{table}

\begin{table*}[ht!]
\resizebox{\textwidth}{!}{
\begin{tabular}{@{}lcccccccccc@{}}
\toprule
\multicolumn{1}{c}{\multirow{2}{*}{System}} & \multicolumn{5}{c}{WMT22}                                                                                             & \multicolumn{5}{c}{FLORES-200}          \\ \cmidrule(lr){2-6} \cmidrule(lr){7-11}
\multicolumn{1}{c}{}                        & \texttt{de}$\rightarrow$\texttt{en}                   & \texttt{en}$\rightarrow$\texttt{de}                     & \texttt{en}$\rightarrow$\texttt{zh}                     & \texttt{zh}$\rightarrow$\texttt{en}                     & Avg.  & \texttt{de}$\rightarrow$\texttt{en} & \texttt{en}$\rightarrow$\texttt{de} & \texttt{en}$\rightarrow$\texttt{zh} & \texttt{zh}$\rightarrow$\texttt{en}& Avg.  \\
\midrule
\multicolumn{11}{@{}c@{}}{\textit{Commercial LLMs \& LLaMA-2-7B based MT systems}} \\
ChatGPT\textsubscript{(3.5-turbo-0613)}   &  85.38   &  86.92       &     87.00   &    82.42   &  85.43  &  89.58   & 88.68      &   88.56    &  86.91     &   88.02    \\
GPT-4\textsubscript{(gpt-4-0613)}  &  85.57   &      87.36    &    87.29       &   82.88 &     85.78  &     89.66    &  88.89     &    88.91   &   87.25    &  88.68     \\
ALMA-7B\textsubscript{(LLaMA-2)}   & 83.98 & 85.59 & 85.05 & 79.73 & 83.59 &     -$^\otimes$   &   -$^\otimes$    &  -$^\otimes$     &   -$^\otimes$    &   -$^\otimes$    \\ \midrule
\multicolumn{11}{@{}c@{}}{\textit{BLOOMZ-mt-7B based LLMs}} \\
ParroT\textsubscript{(BLOOMZ-mt)}                           & 78.00                     & 73.60                     & 83.50                     & 79.00                     & 78.53 & -$^\ast$       & -$^\ast$     & -$^\ast$     & -$^\ast$     & -$^\ast$     \\
TIM\textsubscript{(BLOOMZ-mt)}                              & 77.65                     & 74.16                     & 84.89                     & 79.50                     & 79.05 & -$^\ast$       & -$^\ast$     & -$^\ast$     & -$^\ast$     & -$^\ast$     \\
SWIE\textsubscript{(BLOOMZ-mt)}                                & 78.80                     & 75.17                    & 84.53                    & 79.15                    & 79.41 & -$^\ast$       & -$^\ast$     & -$^\ast$     & -$^\ast$     & -$^\ast$     \\
\noalign{\vskip 0.2ex}\hdashline\noalign{\vskip 0.2ex}
\multicolumn{11}{@{}c@{}}{\textit{LLaMA-1-7B based LLMs}} \\
ParroT\textsubscript{(LLaMA-1)}   & 82.40   & 81.60     & 80.30     & 75.90      & 80.05 & 88.40   & 84.60 & 81.20 & 83.40 & 84.40 \\
TIM\textsubscript{(LLaMA-1)}  & 82.80  & 82.32    & 80.03    & 75.46         & 80.15 & 88.08   & 85.00 & 80.93 & 83.18 & 84.30 \\
SWIE\textsubscript{(LLaMA-1)}       & 82.97       & 81.89        & 80.14                    & 76.14           & 80.29  & 88.39       & 85.21    & 81.14     & 83.50     & 84.56     \\
VicunaMT\textsubscript{(LLaMA-1)}               & 83.55                     & 82.79                     & 81.27                     & 77.39                     & 81.25 &         88.66                        &          86.27                       &                       80.62                                                &                    84.44             &               85.00   \\
+ REF                                      &            83.88               &       83.37                    &         82.86                  &           78.19               &   82.07    &    88.48     &          86.11   &   83.35    &  84.54 & 85.62    \\
+ BEST                                       &        83.61                   &      83.08                     &            83.20               &           78.35                &    82.06   &   88.67      &   85.87    &   84.02    &    84.55   &   85.78    \\
+ PL                                       &           \textbf{84.23}                &      \textbf{84.43 }                    &             \textbf{84.26}              &           \textbf{79.07}                &    \textbf{83.00}   &   \textbf{88.83}      &   \textbf{86.73}    &   \textbf{84.88}    &    \textbf{84.76}   &  \textbf{86.30}     \\ 
\bottomrule
\end{tabular}
}
\caption{Model performance in COMET scores. Best results of LLaMA-1 based models are in \textbf{bold}. Applying prefrence learning (+PL) on top of our VicunaMT model consistently leads to improvements in all cases, achieving the highest average performance among all BLOOM and LLaMA-1 based MT models. Performance in BLEU scores is reported in Appendix~\ref{sec:appx:model_comparison_bleu_score}. $^\otimes$: LLaMA-2 based models were not evaluated due to license constraints. WMT22 results are extracted from the original paper. $^\ast$: BLOOMZ-family models use FLORES-200 for training.}
\label{tab:preference_learning_results}
\end{table*}

\paragraph{Results.} Table \ref{tab:sft-performance-wmt22-comet} presents the results before and after SFT. It can be seen that LLMs without instruction-tuning, e.g., BLOOM, perform poorly; we observe that they tend to overgenerate and repeat tokens in the source sentences.\footnote{Overgeneration is also noticed in~\cite{Bawden2023_bloom_mt}, while it can be partially alleviated by prompt engineering and text post-processing~\cite{Srivastava2023_prompt_enhance_zero_shot}, enhancing LLMs' zero-shot performance is not our primary focus.} In contrast, instruction-tuned models work out-of-the-box and exhibit decent performance. It can be also observed that SFT dramatically boosts the performance of raw LLMs, and slightly benefits instruction-tuned LLMs. For BLOOM and Mistral, the performance gap between raw and instruction-tuned models is mostly lost after SFT. An interesting case is Vicuna, where there is a considerable improvement on \texttt{en}$\leftrightarrow$\texttt{zh} over its base model LLaMA-1. This implies that instruction-tuned LLMs may serve as a better base model for SFT. 
In addition, different LLMs excel in diverse translation directions and their instruction-tuned versions do not deviate from this pattern. For example, both BLOOM and BLOOMZ perform quite well on \texttt{en}$\rightarrow$\texttt{zh}, but have a deficiency in \texttt{en}$\rightarrow$\texttt{de}. For LLaMA-based models, the opposite holds. This could be due to the fact that German and Chinese are not included (at least, not intentionally) in BLOOM's and LLaMA's pre-training corpora, respectively.%

The Vicuna+SFT model has the best overall performance and so we select it as our target LLM to be improved through preference learning. We call this model \textbf{VicunaMT}. The generated translations in the \mtdata dataset are produced by this model.

\subsection{Refining through preference learning}
\label{sec: preference_learning_improves_model_performance}

\paragraph{Baselines.} We continue training our VicunaMT model on \mtdata through preference learning and compare it with the following  competitive systems from recent work: (1) \textbf{ParroT} \cite{Jiao2023_parrot} adds a ``Hint'' field to the model input, prompting the model to generate both correct and incorrect translations. At inference time, the ``correct'' version of the translations is used for evaluation. (2) \textbf{TIM} \cite{Zeng2023_tim} incorporates standard SFT with a ranking loss computed on a pair of correct and incorrect translations. (3) \textbf{SWIE} \cite{Chen2023_SWIE} proposes to attach an instruction adapter to enhance LLMs' long-term attention for better translation. (4) \textbf{ALMA} \cite{Xu2023_alma} first continues pre-training the LLM on monolingual data, followed by performing SFT on parallel data. Furthermore, as the preference learning stage introduces additional data, a performance gain could be trivial by exposing the model with more samples. To establish a fair comparison, we design two additional baselines: (5) \textbf{REF} trains VicunaMT with the reference translations in \mtdata. (6) \textbf{BEST} trains VicunaMT with the translations that are scored highest by our annotators. See Table~\ref{tab:model-better-than-reference} for an example comparison of the reference and best translations. All aforementioned baselines are performed on 7B LLMs (based either on BLOOM-7B or LLaMA-7B). Finally, we also compare our model against commercial LLMs, including ChatGPT and GPT-4.

\paragraph{Results.} We report the MT performance of various baselines in Table \ref{tab:preference_learning_results}. It can be seen that our VicunaMT model performs well compared to recent MT systems. PL further increases the performance advantage. Our final model, VicunaMT+PL, achieves the highest average performance (83 on WMT22 and 86.3 on FLORES-200), consistently outperforming all LLaMA-1 based models across all directions, with the largest improvement being a 3.96 increase in COMET score. (\texttt{en}$\rightarrow$\texttt{zh} on WMT22). Notably, LLaMA-based models are originally much weaker in directions involving Chinese. Through preference learning, VicunaMT reaches a translation performance close to BLOOM-based LLMs. This becomes practically significant when the goal is to deploy a single LLM to handle multiple translation directions. Also, the PL model scores higher than VicunaMT models fine-tuned on the reference and best translations, indicating that the performance gain does not just come from having more data. Compared to the ALMA model, which is based on LLaMA-2~\cite{Touvron2023_llama2}, a widely recognized superior open access LLM, our model demonstrates only a slight deficit of 0.59 COMET scores. Note that our strategy is orthogonal to ALMA's approach, which leverages monolingual data. Combining both strategies should lead to even better performance.

We supplement our assessment with a human evaluation, contrasting VicunaMT+PL with SFT-only Vicuna variations including VicunaMT and VicunaMT+REF, as illustrated in Table~\ref{tab:humaneval}. The human evaluation confirms the trend observed with automatic metrics, where PL substantially outperforms SFT-only variations.

\begin{table}[h]
\centering
\resizebox{\columnwidth}{!}{
\begin{tabular}{@{}lccccc@{}}
\toprule
& \texttt{de}$\rightarrow$\texttt{en}                   & \texttt{en}$\rightarrow$\texttt{de}                     & \texttt{en}$\rightarrow$\texttt{zh}                     & \texttt{zh}$\rightarrow$\texttt{en} \\ \midrule
& \multicolumn{5}{@{}c@{}}{VicunaMT+PL vs.} \\
VicunaMT &    $+3.7\% $      & $+4.4\%$    & $+5.6\% $ & $+5.7\% $  \\
VicunaMT+REF &    $+3.7\% $      & $+2.5\%$   & $+5.0\% $ & $+3.5\% $  \\\bottomrule
\end{tabular}} 
\caption{Relative improvements of VicunaMT+PL over SFT-only models (VicunaMT and VicunaMT+REF), assessed through human evaluation on the WMT22 test set, employing the same scoring criteria as those specified in \mtdata. A two-sided t-test was conducted, with 95\% confidence intervals noted as $\pm 1.7\%$. Positive values indicate the improvement achieved by VicunaMT+PL compared to the other models.}\label{tab:humaneval}
\end{table}

\section{Analysis}
\label{sec:analysis}

\paragraph{Reuse of preference data.}\mtdata contains the translations generated by VicunaMT, which is also the target LLM we aim to improve. There would be additional value if this data could be reused to improve other LLMs. To investigate this, we train both Mistral-Instruct and BLOOMZ on \mtdata using PL. As shown in Table~\ref{tab:preference-data-help-other-LLMs}, PL improves both models, suggesting that the \mtdata is not limited for use with VicunaMT and can be reused for improving other LLMs.

\begin{table}[h]
\resizebox{\columnwidth}{!}{
\begin{tabular}{@{}lccccc@{}}
\toprule
                         & \multicolumn{5}{c}{\textit{WMT22}} \\
                         & \multicolumn{1}{c}{\texttt{de}$\rightarrow$\texttt{en}} & \multicolumn{1}{c}{\texttt{en}$\rightarrow$\texttt{de}} & \texttt{en}$\rightarrow$\texttt{zh}                     & \multicolumn{1}{c}{\texttt{zh}$\rightarrow$\texttt{en}} & Avg.  \\ \midrule
                         
BLOOMZ$^\dagger$ & 77.24  & 69.32 & 84.95    & 78.77       & 77.57 \\
+REF & 77.41  & 68.47 & 84.76   & 79.50      & 77.53 \\
+BEST & 77.48  & 68.64 & 85.15    &79.59      & 77.72 \\
+PL & \textbf{77.83}  & \textbf{69.84} & \textbf{85.36}    & \textbf{80.67}       & \textbf{78.42} \\
\midrule
Mistral-Ins.$^\dagger$ & 82.68    & 81.23     & 82.49 & 77.73       & 81.03 \\
+REF & 83.06    & 82.63     & 83.39 & 78.07       & 81.79 \\
+BEST & 82.98    & 81.84     & 83.34 & 78.33       & 81.62 \\
+PL & \textbf{83.35}    & \textbf{82.94}     & \textbf{84.71} & \textbf{79.25}       & \textbf{82.56} \\
\bottomrule
\end{tabular}
}
\caption{Model performance in COMET scores. Best results are in \textbf{bold}. \mtdata can be reused to improve BLOOMZ and Mistral-Instruct. See results on FLORES-200 and in BLEU scores in Appendix~\ref{sec:appx:data_reuse_bleu_score}.$^\dagger$: SFT stage has already been applied to these models.}
\label{tab:preference-data-help-other-LLMs}
\end{table}

\begin{figure}[h]
    \centering
    \includegraphics[width=\columnwidth]{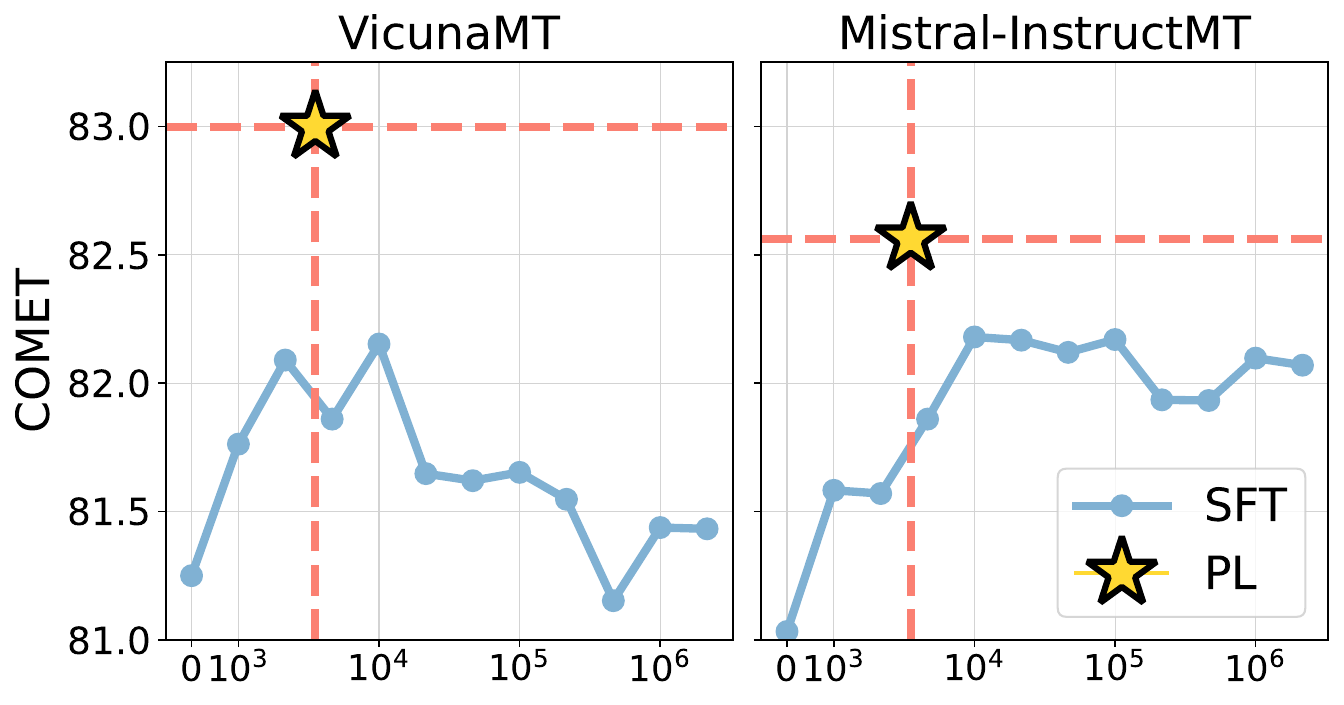}
    \caption{Performance comparison between PL using 4.4K examples from \mtdata and SFT, employing up to 1.4M parallel data. Evaluation is done on WMT22, and COMET scores are averaged across four translation directions. Performing SFT on more parallel data does not always lead to performance gain. PL consistently outperforms SFT in all cases.}
    \label{fig:limited_gain_parallel_data}
\end{figure}

\paragraph{Limited gains with additional parallel data.} Section~\ref{sec: preference_learning_improves_model_performance} shows that the \mtdata dataset, which contains 4.4K preference examples, can be more valuable than an equivalent amount of parallel data with either the reference or the best translations. A natural follow-up question is whether adding more parallel data can close the gap. To answer this question, we collect more data by concatenating WMT20, WMT21 test data with News Commentary v16, making 1.4M parallel sentences in total.\footnote{We select News Commentary for its high-quality, domain-matching parallel data to WMT test data. WMT20/21 are included as \mtdata is built on a subset from them.} We fine-tune VicunaMT and Mistral-InstructMT (i.e., Mistral-Instruct after SFT stage) on different proportions of this data and plot the performance curve in Figure~\ref{fig:limited_gain_parallel_data}. In both cases, similar to observations in \cite{Xu2023_alma}, adding more parallel data does not always improve these models and they never attain the performance level reached by using PL with \mtdata.

\begin{figure}[h]
    \centering
    \includegraphics[width=\columnwidth]{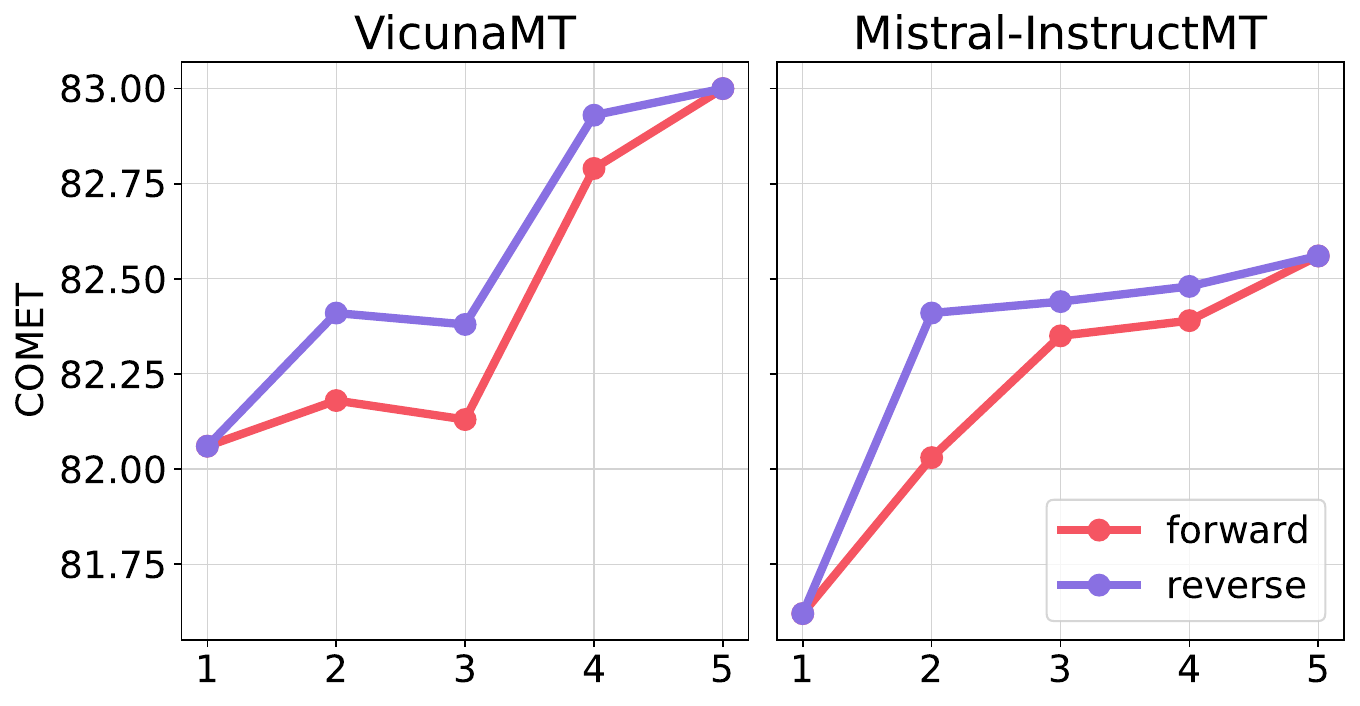}
    \caption{Model performance varying number of translations ($K$) per source sentence. Evaluation conducted on WMT22 and COMET scores averaged across four translation directions are reported. Reverse mode selects more diverse translations and achieves better performance, especially when fewer translations are provided.
    }
    \label{fig:forward_reverse_comparison}
\end{figure}

\paragraph{Diverse translations help more.}
By default, we perform PL using all five translations provided by \mtdata. We now study the relation between the final model performance and the number of preference translations used. We select $K=\{2,3,4\}$ translations and rerun the PL algorithm on VicunaMT and Mistral-InstructMT. We explore two selection modes for selecting $K$ translations. Given five translations sorted by human preference scores in descending order, the \textit{forward} mode selects the first $K$ translations (i.e., the best $K$), while the \textit{reverse} mode select the first and last $K-1$ translations. We compare both modes varying $K$ and present the results in Figure~\ref{fig:forward_reverse_comparison}. There is a clear disparity in performance with these two selection modes. The reverse mode consistently outperforms the forward mode given the same number of translations, with a larger advantage in low-resource cases, such as when $K=2$. This is intuitive since the reverse mode always includes the highest- and lowest-scored translations and thus, PL may have a better chance to see ``hard negatives'' which have low human preference score but high generation probability. The general trend shows that including more preference samples is better, and using all available samples yields the best performance.

\begin{table}[h]
\resizebox{\columnwidth}{!}{
\begin{tabular}{@{}lcc@{}}
\toprule
          & VicunaMT & Mistral-InstrctMT \\ \midrule
SFT stage            & 81.25    & 81.03            \\
\noalign{\vskip 0.2ex}\hdashline\noalign{\vskip 0.2ex}

PL  stage             & 83.00    & 82.56             \\
\ \ w/o $\mathcal{L}_{SFT}$          & 83.00    & 82.54             \\
\ \ w/o distance          & 82.22    & 81.92           \\
\ \ w/o $\mathcal{L}_{SFT}$/dist. & 74.65    & 60.70             \\ 
\ \ $\mathcal{L}_{SFT}$ only              & 82.07    & 81.79            \\
\bottomrule
\end{tabular}
}
\caption{Ablation study. PL is less sensitive to $\mathcal{L}_{SFT}$ than the distance information. Disabling both factors leads to substantial model degradation.}
\label{tab:pl-ablation-study}
\end{table}

\paragraph{Distance information is crucial.} Our framework considers the distance information in preference scores (Equation~\ref{eq:plackett-luce-loss-with-distance}). We now investigate if this information can be replaced by simply using the ranking information. That is, we set $d_{i}^{j}=1$ for all translations and rerun the PL algorithm. Table~\ref{tab:pl-ablation-study} shows that when the distance information is available, excluding the SFT loss does not harm the performance much. In fact, we achieve the best performance when setting $\beta=0$ for VicunaMT. However, when the distance information is withheld, we see a clear degradation in performance. We find that a larger $\beta$ value is required when relying only on the ranking information, but this makes the PL algorithm closer to SFT. As a result, when only the ranking information is provided, VicunaMT performs similarly to the $\mathcal{L}_{SFT}$ only baseline. Finally, disabling both $L_{SFT}$ and distance cause a large performance drop.
\begin{figure}[h]
    \centering
    \includegraphics[width=\columnwidth]{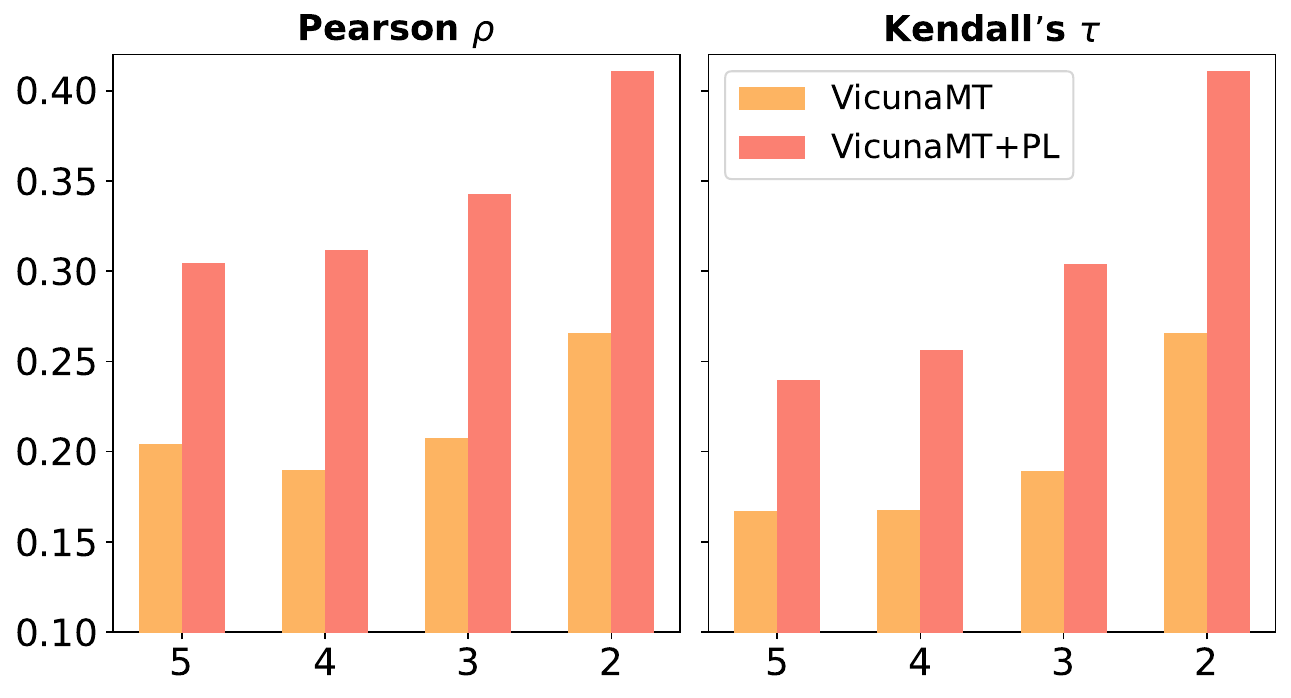}
    \caption{Sentence-level correlation between model generation probability and human preference scores varying number of translations ($K$). PL helps the model align better with human judgement.}
    \label{fig:model_calibration}
\end{figure}
\paragraph{Better Model Calibration.} In our preference learning framework, the model learns both translation and the ability to differentiate between different translation quality. We analyze if PL has successfully transferred human preference to the model. Using the held-out set of \mtdata, we examine the sentence-level correlation between the scores assigned by the human annotators and model generation probability. Specifically, we compute the average Pearson and Kendall's tau correlation varying the number of preference samples (reverse mode). The results are presented in Figure~\ref{fig:model_calibration}. Compared to the SFT baseline, VicunaMT, PL substantially improves the correlation, suggesting that our final model aligns better with human preference.

\section{Conclusion} We present a preference learning framework to break the performance plateau faced when performing SFT. It enhances the translation capabilities of LLMs by motivating them to differentiate the nuances in different translations. To support this framework, we have carefully curated a preference dataset, named \mtdata, featuring translations of varying quality, each scored by professional translators. Extensive experiments, including human evaluations, confirm the effectiveness of this framework. In addition, we demonstrate that \mtdata can be reused to enhance other LLMs, further bolstering its practical usability. Future research could consider extending our framework into an iterative process for continuous improvement of LLMs' translation capabilities.

\section*{Limitations}
This work demonstrates that preference learning can effectively improve LLMs' translation capabilities. However, our
study is not exhaustive and has the following limitations.

\paragraph{Low-resource languages.} This work centers on translation directions involving high-resource languages where LLMs already exhibit proficiency. The extent to which translations for low-resource languages can leverage our framework remains uncertain. Nevertheless, it is important to emphasize that our framework is language- and model-agnostic, implying its potential applicability to low-resource languages. We leave the investigation into this aspect to future work.

\paragraph{Annotation cost.} Assigning preference scores for five translations per sentence can be costly, which may hinder the scaling up of the preference dataset. However, as we show in Section~\ref{sec:analysis}, a preference learning dataset such as \mtdata offers distinct learning signal that is not covered by massive parallel data. In addition, we highlight that the preference data can be reused to benefit other LLMs. Thus, the collected data is a valuable and reusable resource, rather than a one-time expense.

\paragraph{Noise in human judgement.} Inevitably, human preference scores can be subjective, and annotators may not always agree. Additionally, there is a risk of annotators finding shortcuts in the annotation process~\cite{Ipeirotis2010_human_short_cut, Hosking2023_human_not_gold}. To reduce the potential annotation mistakes, we average the scores of two translators for each sample and all translators we employ are experienced in translation assessment. 

\section*{Acknowledgements}

Bill Byrne holds concurrent appointments as an Amazon Scholar and as Professor of Information Engineering at the University of Cambridge. This paper describes work performed at Amazon.

We thank Felix Hieber, Lei Sun, and Tobias Domhan for their thoughtful advice, in-depth discussions, and implementation support. We would also like to thank our anonymous reviewers for their constructive feedback.

\bibliography{preference_mt}

\begin{thebibliography}{61}
\expandafter\ifx\csname natexlab\endcsname\relax\def\natexlab#1{#1}\fi

\bibitem[{Alves et~al.(2023)Alves, Guerreiro, Alves, Pombal, Rei, de~Souza, Colombo, and Martins}]{Alves2023}
Duarte~M. Alves, Nuno~Miguel Guerreiro, Jo{\~{a}}o Alves, Jos{\'{e}} Pombal, Ricardo Rei, Jos{\'{e}} G.~C. de~Souza, Pierre Colombo, and Andr{\'{e}} F.~T. Martins. 2023.
\newblock \href {https://doi.org/10.48550/ARXIV.2310.13448} {Steering large language models for machine translation with finetuning and in-context learning}.
\newblock \emph{CoRR}, abs/2310.13448.

\bibitem[{Bawden and Yvon(2023)}]{Bawden2023_bloom_mt}
Rachel Bawden and Fran{\c{c}}ois Yvon. 2023.
\newblock \href {https://aclanthology.org/2023.eamt-1.16} {Investigating the translation performance of a large multilingual language model: the case of {BLOOM}}.
\newblock In \emph{Proceedings of the 24th Annual Conference of the European Association for Machine Translation, {EAMT} 2023, Tampere, Finland, 12-15 June 2023}, pages 157--170. European Association for Machine Translation.

\bibitem[{Bradley(1953)}]{Bradley1953}
Ralph~Allan Bradley. 1953.
\newblock \href {https://doi.org/10.2307/3001630} {Some statistical methods in taste testing and quality evaluation}.
\newblock \emph{Biometrics}, 9(1):22--38.

\bibitem[{Bradley and Terry(1952)}]{Bradley1952}
Ralph~Allan Bradley and Milton~E Terry. 1952.
\newblock Rank analysis of incomplete block designs: I. the method of paired comparisons.
\newblock \emph{Biometrika}, 39(3/4):324--345.

\bibitem[{Brown et~al.(2020)Brown, Mann, Ryder, Subbiah, Kaplan, Dhariwal, Neelakantan, Shyam, Sastry, Askell et~al.}]{Brown2020_gpt3}
Tom Brown, Benjamin Mann, Nick Ryder, Melanie Subbiah, Jared~D Kaplan, Prafulla Dhariwal, Arvind Neelakantan, Pranav Shyam, Girish Sastry, Amanda Askell, et~al. 2020.
\newblock Language models are few-shot learners.
\newblock \emph{Advances in neural information processing systems}, 33:1877--1901.

\bibitem[{Chen et~al.(2023)Chen, Liu, Meng, Chen, Xu, and Zhou}]{Chen2023_SWIE}
Yijie Chen, Yijin Liu, Fandong Meng, Yufeng Chen, Jinan Xu, and Jie Zhou. 2023.
\newblock Improving translation faithfulness of large language models via augmenting instructions.
\newblock \emph{arXiv preprint arXiv:2308.12674}.

\bibitem[{Cheng et~al.(2010)Cheng, Dembczynski, and H{\"{u}}llermeier}]{Cheng2010}
Weiwei Cheng, Krzysztof Dembczynski, and Eyke H{\"{u}}llermeier. 2010.
\newblock \href {https://icml.cc/Conferences/2010/papers/353.pdf} {Label ranking methods based on the plackett-luce model}.
\newblock In \emph{Proceedings of the 27th International Conference on Machine Learning (ICML-10), June 21-24, 2010, Haifa, Israel}, pages 215--222. Omnipress.

\bibitem[{Cheng and H\"ullermeier(2008)}]{Cheng2008}
Weiwei Cheng and Eyke H\"ullermeier. 2008.
\newblock Learning similarity functions from qualitative feedback.
\newblock In \emph{LNAI 5239 Advances in Case-Based Reasoning: The 9th European Conference on Case-Based Reasoning (ECCBR-08)}, pages 129--134, Trier, Germany. Springer.

\bibitem[{Chowdhery et~al.(2023)Chowdhery, Narang, Devlin, Bosma, Mishra, Roberts, Barham, Chung, Sutton, Gehrmann, Schuh, Shi, Tsvyashchenko, Maynez, Rao, Barnes, Tay, Shazeer, Prabhakaran, Reif, Du, Hutchinson, Pope, Bradbury, Austin, Isard, Gur-Ari, Yin, Duke, Levskaya, Ghemawat, Dev, Michalewski, Garcia, Misra, Robinson, Fedus, Zhou, Ippolito, Luan, Lim, Zoph, Spiridonov, Sepassi, Dohan, Agrawal, Omernick, Dai, Pillai, Pellat, Lewkowycz, Moreira, Child, Polozov, Lee, Zhou, Wang, Saeta, Diaz, Firat, Catasta, Wei, Meier-Hellstern, Eck, Dean, Petrov, and Fiedel}]{Chowdhery2023_palm}
Aakanksha Chowdhery, Sharan Narang, Jacob Devlin, Maarten Bosma, Gaurav Mishra, Adam Roberts, Paul Barham, Hyung~Won Chung, Charles Sutton, Sebastian Gehrmann, Parker Schuh, Kensen Shi, Sasha Tsvyashchenko, Joshua Maynez, Abhishek Rao, Parker Barnes, Yi~Tay, Noam Shazeer, Vinodkumar Prabhakaran, Emily Reif, Nan Du, Ben Hutchinson, Reiner Pope, James Bradbury, Jacob Austin, Michael Isard, Guy Gur-Ari, Pengcheng Yin, Toju Duke, Anselm Levskaya, Sanjay Ghemawat, Sunipa Dev, Henryk Michalewski, Xavier Garcia, Vedant Misra, Kevin Robinson, Liam Fedus, Denny Zhou, Daphne Ippolito, David Luan, Hyeontaek Lim, Barret Zoph, Alexander Spiridonov, Ryan Sepassi, David Dohan, Shivani Agrawal, Mark Omernick, Andrew~M. Dai, Thanumalayan~Sankaranarayana Pillai, Marie Pellat, Aitor Lewkowycz, Erica Moreira, Rewon Child, Oleksandr Polozov, Katherine Lee, Zongwei Zhou, Xuezhi Wang, Brennan Saeta, Mark Diaz, Orhan Firat, Michele Catasta, Jason Wei, Kathy Meier-Hellstern, Douglas Eck, Jeff Dean, Slav Petrov, and Noah Fiedel. 2023.
\newblock \href {http://jmlr.org/papers/v24/22-1144.html} {Palm: Scaling language modeling with pathways}.
\newblock \emph{Journal of Machine Learning Research}, 24(240):1--113.

\bibitem[{Costa{-}juss{\`{a}} et~al.(2022)Costa{-}juss{\`{a}}, Cross, {\c{C}}elebi, Elbayad, Heafield, Heffernan, Kalbassi, Lam, Licht, Maillard, Sun, Wang, Wenzek, Youngblood, Akula, Barrault, Gonzalez, Hansanti, Hoffman, Jarrett, Sadagopan, Rowe, Spruit, Tran, Andrews, Ayan, Bhosale, Edunov, Fan, Gao, Goswami, Guzm{\'{a}}n, Koehn, Mourachko, Ropers, Saleem, Schwenk, and Wang}]{Costajussa2022_nllb}
Marta~R. Costa{-}juss{\`{a}}, James Cross, Onur {\c{C}}elebi, Maha Elbayad, Kenneth Heafield, Kevin Heffernan, Elahe Kalbassi, Janice Lam, Daniel Licht, Jean Maillard, Anna Sun, Skyler Wang, Guillaume Wenzek, Al~Youngblood, Bapi Akula, Lo{\"{\i}}c Barrault, Gabriel~Mejia Gonzalez, Prangthip Hansanti, John Hoffman, Semarley Jarrett, Kaushik~Ram Sadagopan, Dirk Rowe, Shannon Spruit, Chau Tran, Pierre Andrews, Necip~Fazil Ayan, Shruti Bhosale, Sergey Edunov, Angela Fan, Cynthia Gao, Vedanuj Goswami, Francisco Guzm{\'{a}}n, Philipp Koehn, Alexandre Mourachko, Christophe Ropers, Safiyyah Saleem, Holger Schwenk, and Jeff Wang. 2022.
\newblock \href {https://doi.org/10.48550/ARXIV.2207.04672} {No language left behind: Scaling human-centered machine translation}.
\newblock \emph{CoRR}, abs/2207.04672.

\bibitem[{Dong et~al.(2023)Dong, Wang, Sreedhar, Wu, and Kuchaiev}]{Dong2023_SteerLM}
Yi~Dong, Zhilin Wang, Makesh~Narsimhan Sreedhar, Xianchao Wu, and Oleksii Kuchaiev. 2023.
\newblock \href {https://doi.org/10.48550/ARXIV.2310.05344} {Steerlm: Attribute conditioned {SFT} as an (user-steerable) alternative to {RLHF}}.
\newblock \emph{CoRR}, abs/2310.05344.

\bibitem[{Freitag et~al.(2022)Freitag, Rei, Mathur, Lo, Stewart, Avramidis, Kocmi, Foster, Lavie, and Martins}]{Freitag2022}
Markus Freitag, Ricardo Rei, Nitika Mathur, Chi-kiu Lo, Craig Stewart, Eleftherios Avramidis, Tom Kocmi, George Foster, Alon Lavie, and Andr{\'e} F.~T. Martins. 2022.
\newblock \href {https://aclanthology.org/2022.wmt-1.2} {Results of {WMT}22 metrics shared task: Stop using {BLEU} {--} neural metrics are better and more robust}.
\newblock In \emph{Proceedings of the Seventh Conference on Machine Translation (WMT)}, pages 46--68, Abu Dhabi, United Arab Emirates (Hybrid). Association for Computational Linguistics.

\bibitem[{Ghazvininejad et~al.(2023)Ghazvininejad, Gonen, and Zettlemoyer}]{Ghazvininejad2023}
Marjan Ghazvininejad, Hila Gonen, and Luke Zettlemoyer. 2023.
\newblock \href {https://doi.org/10.48550/ARXIV.2302.07856} {Dictionary-based phrase-level prompting of large language models for machine translation}.
\newblock \emph{CoRR}, abs/2302.07856.

\bibitem[{Hamilton et~al.(2023)Hamilton, Tawn, and Firth}]{hamilton_many_2023}
Ian Hamilton, Nick Tawn, and David Firth. 2023.
\newblock \href {https://doi.org/10.48550/arXiv.2312.13619} {The many routes to the ubiquitous {B}radley-{T}erry model}.
\newblock ({arXiv}:2312.13619).

\bibitem[{He et~al.(2023)He, Liang, Jiao, Zhang, Yang, Wang, Tu, Shi, and Wang}]{He2023}
Zhiwei He, Tian Liang, Wenxiang Jiao, Zhuosheng Zhang, Yujiu Yang, Rui Wang, Zhaopeng Tu, Shuming Shi, and Xing Wang. 2023.
\newblock \href {https://doi.org/10.48550/ARXIV.2305.04118} {Exploring human-like translation strategy with large language models}.
\newblock \emph{CoRR}, abs/2305.04118.

\bibitem[{Hejna et~al.(2023)Hejna, Rafailov, Sikchi, Finn, Niekum, Knox, and Sadigh}]{Hejna2023_cpl}
Joey Hejna, Rafael Rafailov, Harshit Sikchi, Chelsea Finn, Scott Niekum, W.~Bradley Knox, and Dorsa Sadigh. 2023.
\newblock \href {https://doi.org/10.48550/ARXIV.2310.13639} {Contrastive preference learning: Learning from human feedback without {RL}}.
\newblock \emph{CoRR}, abs/2310.13639.

\bibitem[{Hendy et~al.(2023)Hendy, Abdelrehim, Sharaf, Raunak, Gabr, Matsushita, Kim, Afify, and Awadalla}]{Hendy2023_chatgpt_translation}
Amr Hendy, Mohamed Abdelrehim, Amr Sharaf, Vikas Raunak, Mohamed Gabr, Hitokazu Matsushita, Young~Jin Kim, Mohamed Afify, and Hany~Hassan Awadalla. 2023.
\newblock How good are {GPT} models at machine translation? a comprehensive evaluation.
\newblock \emph{arXiv preprint arXiv:2302.09210}.

\bibitem[{Herold et~al.(2022)Herold, Rosendahl, Vanvinckenroye, and Ney}]{Herold2022}
Christian Herold, Jan Rosendahl, Joris Vanvinckenroye, and Hermann Ney. 2022.
\newblock \href {https://doi.org/10.18653/V1/2022.FINDINGS-ACL.200} {Detecting various types of noise for neural machine translation}.
\newblock In \emph{Findings of the Association for Computational Linguistics: {ACL} 2022, Dublin, Ireland, May 22-27, 2022}, pages 2542--2551. Association for Computational Linguistics.

\bibitem[{Holtzman et~al.(2020)Holtzman, Buys, Du, Forbes, and Choi}]{Holtzman2020}
Ari Holtzman, Jan Buys, Li~Du, Maxwell Forbes, and Yejin Choi. 2020.
\newblock \href {https://openreview.net/forum?id=rygGQyrFvH} {The curious case of neural text degeneration}.
\newblock In \emph{8th International Conference on Learning Representations, {ICLR} 2020, Addis Ababa, Ethiopia, April 26-30, 2020}. OpenReview.net.

\bibitem[{Hosking et~al.(2023)Hosking, Blunsom, and Bartolo}]{Hosking2023_human_not_gold}
Tom Hosking, Phil Blunsom, and Max Bartolo. 2023.
\newblock \href {https://doi.org/10.48550/ARXIV.2309.16349} {Human feedback is not gold standard}.
\newblock \emph{CoRR}, abs/2309.16349.

\bibitem[{Hu et~al.(2023)Hu, Tao, Yang, and Zhou}]{Hu2023_align_llm_offline}
Jian Hu, Li~Tao, June Yang, and Chandler Zhou. 2023.
\newblock \href {https://doi.org/10.48550/ARXIV.2308.12050} {Aligning language models with offline reinforcement learning from human feedback}.
\newblock \emph{CoRR}, abs/2308.12050.

\bibitem[{Huang et~al.(2022)Huang, Dossa, Raffin, Kanervisto, and Wang}]{Huang2022}
Shengyi Huang, Rousslan Fernand~Julien Dossa, Antonin Raffin, Anssi Kanervisto, and Weixun Wang. 2022.
\newblock \href {https://iclr-blog-track.github.io/2022/03/25/ppo-implementation-details/} {The 37 implementation details of proximal policy optimization}.
\newblock In \emph{ICLR Blog Track}.
\newblock Https://iclr-blog-track.github.io/2022/03/25/ppo-implementation-details/.

\bibitem[{Ipeirotis et~al.(2010)Ipeirotis, Provost, and Wang}]{Ipeirotis2010_human_short_cut}
Panagiotis~G Ipeirotis, Foster Provost, and Jing Wang. 2010.
\newblock Quality management on {A}mazon {M}echanical {T}urk.
\newblock In \emph{Proceedings of the ACM SIGKDD workshop on human computation}, pages 64--67.

\bibitem[{Islam et~al.(2017)Islam, Henderson, Gomrokchi, and Precup}]{Islam2017}
Riashat Islam, Peter Henderson, Maziar Gomrokchi, and Doina Precup. 2017.
\newblock \href {http://arxiv.org/abs/1708.04133} {Reproducibility of benchmarked deep reinforcement learning tasks for continuous control}.
\newblock \emph{CoRR}, abs/1708.04133.

\bibitem[{Jiang et~al.(2023)Jiang, Sablayrolles, Mensch, Bamford, Chaplot, de~Las~Casas, Bressand, Lengyel, Lample, Saulnier, Lavaud, Lachaux, Stock, Scao, Lavril, Wang, Lacroix, and Sayed}]{Jiang2023_mistral}
Albert~Q. Jiang, Alexandre Sablayrolles, Arthur Mensch, Chris Bamford, Devendra~Singh Chaplot, Diego de~Las~Casas, Florian Bressand, Gianna Lengyel, Guillaume Lample, Lucile Saulnier, L{\'{e}}lio~Renard Lavaud, Marie{-}Anne Lachaux, Pierre Stock, Teven~Le Scao, Thibaut Lavril, Thomas Wang, Timoth{\'{e}}e Lacroix, and William~El Sayed. 2023.
\newblock \href {https://doi.org/10.48550/ARXIV.2310.06825} {Mistral 7b}.
\newblock \emph{CoRR}, abs/2310.06825.

\bibitem[{Jiao et~al.(2023{\natexlab{a}})Jiao, Huang, Wang, Wang, Shi, and Tu}]{Jiao2023_parrot}
Wenxiang Jiao, Jen-tse Huang, Wenxuan Wang, Xing Wang, Shuming Shi, and Zhaopeng Tu. 2023{\natexlab{a}}.
\newblock Parrot: Translating during chat using large language models.
\newblock \emph{arXiv preprint arXiv:2304.02426}.

\bibitem[{Jiao et~al.(2023{\natexlab{b}})Jiao, Wang, Huang, Wang, and Tu}]{Jiao2023_gpt4_rivals_google_deepl}
Wenxiang Jiao, Wenxuan Wang, JT~Huang, Xing Wang, and ZP~Tu. 2023{\natexlab{b}}.
\newblock Is {ChatGPT} a good translator? yes with {GPT-4} as the engine.
\newblock \emph{arXiv preprint arXiv:2301.08745}.

\bibitem[{Khayrallah and Koehn(2018)}]{Khayrallah2018_translation_noise}
Huda Khayrallah and Philipp Koehn. 2018.
\newblock \href {https://doi.org/10.18653/V1/W18-2709} {On the impact of various types of noise on neural machine translation}.
\newblock In \emph{Proceedings of the 2nd Workshop on Neural Machine Translation and Generation, NMT@ACL 2018, Melbourne, Australia, July 20, 2018}, pages 74--83. Association for Computational Linguistics.

\bibitem[{Kocmi et~al.(2022)Kocmi, Bawden, Bojar, Dvorkovich, Federmann, Fishel, Gowda, Graham, Grundkiewicz, Haddow, Knowles, Koehn, Monz, Morishita, Nagata, Nakazawa, Nov{\'a}k, Popel, and Popovi{\'c}}]{Kocmi2022_wmt22}
Tom Kocmi, Rachel Bawden, Ond{\v{r}}ej Bojar, Anton Dvorkovich, Christian Federmann, Mark Fishel, Thamme Gowda, Yvette Graham, Roman Grundkiewicz, Barry Haddow, Rebecca Knowles, Philipp Koehn, Christof Monz, Makoto Morishita, Masaaki Nagata, Toshiaki Nakazawa, Michal Nov{\'a}k, Martin Popel, and Maja Popovi{\'c}. 2022.
\newblock \href {https://aclanthology.org/2022.wmt-1.1} {Findings of the 2022 conference on machine translation ({WMT}22)}.
\newblock In \emph{Proceedings of the Seventh Conference on Machine Translation (WMT)}, pages 1--45, Abu Dhabi, United Arab Emirates (Hybrid). Association for Computational Linguistics.

\bibitem[{Luce(1959)}]{luce_individual_choice_behaviour_1959}
R.~Duncan Luce. 1959.
\newblock \emph{Individual choice behaviour}.
\newblock John Wiley.

\bibitem[{Maillard et~al.(2023)Maillard, Gao, Kalbassi, Sadagopan, Goswami, Koehn, Fan, and Guzman}]{Maillard2023}
Jean Maillard, Cynthia Gao, Elahe Kalbassi, Kaushik~Ram Sadagopan, Vedanuj Goswami, Philipp Koehn, Angela Fan, and Francisco Guzman. 2023.
\newblock \href {https://doi.org/10.18653/v1/2023.acl-long.154} {Small data, big impact: Leveraging minimal data for effective machine translation}.
\newblock In \emph{Proceedings of the 61st Annual Meeting of the Association for Computational Linguistics (Volume 1: Long Papers)}, pages 2740--2756, Toronto, Canada. Association for Computational Linguistics.

\bibitem[{Maystre and Grossglauser(2015)}]{maystre_fast_2015}
Lucas Maystre and Matthias Grossglauser. 2015.
\newblock \href {https://proceedings.neurips.cc/paper_files/paper/2015/hash/2a38a4a9316c49e5a833517c45d31070-Abstract.html} {Fast and accurate inference of {P}lackett-{L}uce models}.
\newblock In \emph{Advances in Neural Information Processing Systems}, volume~28.

\bibitem[{Mosteller(1951)}]{Mosteller1951}
Frederick Mosteller. 1951.
\newblock \href {https://doi.org/10.1007/BF02313422} {Remarks on the method of paired comparisons: I. the least squares solution assuming equal standard deviations and equal correlations}.
\newblock \emph{Psychometrika}, 16(1):3--9.

\bibitem[{Mu et~al.(2023)Mu, Reheman, Cao, Fan, Li, Li, Xiao, Zhang, and Zhu}]{Mu2023}
Yongyu Mu, Abudurexiti Reheman, Zhiquan Cao, Yuchun Fan, Bei Li, Yinqiao Li, Tong Xiao, Chunliang Zhang, and Jingbo Zhu. 2023.
\newblock \href {https://doi.org/10.18653/v1/2023.findings-acl.653} {Augmenting large language model translators via translation memories}.
\newblock In \emph{Findings of the Association for Computational Linguistics: ACL 2023}, pages 10287--10299, Toronto, Canada. Association for Computational Linguistics.

\bibitem[{Muennighoff et~al.(2023)Muennighoff, Wang, Sutawika, Roberts, Biderman, Scao, Bari, Shen, Yong, Schoelkopf, Tang, Radev, Aji, Almubarak, Albanie, Alyafeai, Webson, Raff, and Raffel}]{Muennighoff2023_bloomz}
Niklas Muennighoff, Thomas Wang, Lintang Sutawika, Adam Roberts, Stella Biderman, Teven~Le Scao, M.~Saiful Bari, Sheng Shen, Zheng~Xin Yong, Hailey Schoelkopf, Xiangru Tang, Dragomir Radev, Alham~Fikri Aji, Khalid Almubarak, Samuel Albanie, Zaid Alyafeai, Albert Webson, Edward Raff, and Colin Raffel. 2023.
\newblock \href {https://doi.org/10.18653/V1/2023.ACL-LONG.891} {Crosslingual generalization through multitask finetuning}.
\newblock In \emph{Proceedings of the 61st Annual Meeting of the Association for Computational Linguistics (Volume 1: Long Papers), {ACL} 2023, Toronto, Canada, July 9-14, 2023}, pages 15991--16111. Association for Computational Linguistics.

\bibitem[{OpenAI(2023)}]{OpenAI2023_gpt4}
OpenAI. 2023.
\newblock \href {https://doi.org/10.48550/ARXIV.2303.08774} {{GPT-4} technical report}.
\newblock \emph{CoRR}, abs/2303.08774.

\bibitem[{Ott et~al.(2018)Ott, Auli, Grangier, and Ranzato}]{Ott2018}
Myle Ott, Michael Auli, David Grangier, and Marc'Aurelio Ranzato. 2018.
\newblock \href {http://proceedings.mlr.press/v80/ott18a.html} {Analyzing uncertainty in neural machine translation}.
\newblock In \emph{Proceedings of the 35th International Conference on Machine Learning, {ICML} 2018, Stockholmsm{\"{a}}ssan, Stockholm, Sweden, July 10-15, 2018}, volume~80 of \emph{Proceedings of Machine Learning Research}, pages 3953--3962. {PMLR}.

\bibitem[{Ouyang et~al.(2022)Ouyang, Wu, Jiang, Almeida, Wainwright, Mishkin, Zhang, Agarwal, Slama, Ray, Schulman, Hilton, Kelton, Miller, Simens, Askell, Welinder, Christiano, Leike, and Lowe}]{Ouyang2022_instructGPT}
Long Ouyang, Jeffrey Wu, Xu~Jiang, Diogo Almeida, Carroll Wainwright, Pamela Mishkin, Chong Zhang, Sandhini Agarwal, Katarina Slama, Alex Ray, John Schulman, Jacob Hilton, Fraser Kelton, Luke Miller, Maddie Simens, Amanda Askell, Peter Welinder, Paul~F Christiano, Jan Leike, and Ryan Lowe. 2022.
\newblock \href {https://proceedings.neurips.cc/paper_files/paper/2022/file/b1efde53be364a73914f58805a001731-Paper-Conference.pdf} {Training language models to follow instructions with human feedback}.
\newblock In \emph{Advances in Neural Information Processing Systems}, volume~35, pages 27730--27744. Curran Associates, Inc.

\bibitem[{Papineni et~al.(2002)Papineni, Roukos, Ward, and Zhu}]{Papineni2002_bleu}
Kishore Papineni, Salim Roukos, Todd Ward, and Wei-Jing Zhu. 2002.
\newblock \href {https://doi.org/10.3115/1073083.1073135} {{B}leu: a method for automatic evaluation of machine translation}.
\newblock In \emph{Proceedings of the 40th Annual Meeting of the Association for Computational Linguistics}, pages 311--318, Philadelphia, Pennsylvania, USA. Association for Computational Linguistics.

\bibitem[{Plackett(1975)}]{Plackett1975_pl_model}
Robin~L Plackett. 1975.
\newblock The analysis of permutations.
\newblock \emph{Journal of the Royal Statistical Society Series C: Applied Statistics}, 24(2):193--202.

\bibitem[{Rafailov et~al.(2023)Rafailov, Sharma, Mitchell, Ermon, Manning, and Finn}]{Rafailov2023_dpo}
Rafael Rafailov, Archit Sharma, Eric Mitchell, Stefano Ermon, Christopher~D. Manning, and Chelsea Finn. 2023.
\newblock \href {https://doi.org/10.48550/ARXIV.2305.18290} {Direct preference optimization: Your language model is secretly a reward model}.
\newblock \emph{CoRR}, abs/2305.18290.

\bibitem[{Rei et~al.(2022)Rei, de~Souza, Alves, Zerva, Farinha, Glushkova, Lavie, Coheur, and Martins}]{Rei2022_comet}
Ricardo Rei, Jos{\'{e}} G.~C. de~Souza, Duarte~M. Alves, Chrysoula Zerva, Ana~C. Farinha, Taisiya Glushkova, Alon Lavie, Lu{\'{\i}}sa Coheur, and Andr{\'{e}} F.~T. Martins. 2022.
\newblock \href {https://aclanthology.org/2022.wmt-1.52} {{COMET-22:} {Unbabel-IST} 2022 submission for the metrics shared task}.
\newblock In \emph{Proceedings of the Seventh Conference on Machine Translation, {WMT} 2022, Abu Dhabi, United Arab Emirates (Hybrid), December 7-8, 2022}, pages 578--585. Association for Computational Linguistics.

\bibitem[{Robinson et~al.(2021)Robinson, Chuang, Sra, and Jegelka}]{Robinson2021}
Joshua~David Robinson, Ching{-}Yao Chuang, Suvrit Sra, and Stefanie Jegelka. 2021.
\newblock \href {https://openreview.net/forum?id=CR1XOQ0UTh-} {Contrastive learning with hard negative samples}.
\newblock In \emph{9th International Conference on Learning Representations, {ICLR} 2021, Virtual Event, Austria, May 3-7, 2021}. OpenReview.net.

\bibitem[{Scao et~al.(2022)Scao, Fan, Akiki, Pavlick, Ilic, Hesslow, Castagn{\'{e}}, Luccioni, Yvon, Gall{\'{e}}, Tow, Rush, Biderman, Webson, Ammanamanchi, Wang, Sagot, Muennighoff, del Moral, Ruwase, Bawden, Bekman, McMillan{-}Major, Beltagy, Nguyen, Saulnier, Tan, Suarez, Sanh, Lauren{\c{c}}on, Jernite, Launay, Mitchell, Raffel, Gokaslan, Simhi, Soroa, Aji, Alfassy, Rogers, Nitzav, Xu, Mou, Emezue, Klamm, Leong, van Strien, Adelani, and et~al.}]{Scao2022_bloom}
Teven~Le Scao, Angela Fan, Christopher Akiki, Ellie Pavlick, Suzana Ilic, Daniel Hesslow, Roman Castagn{\'{e}}, Alexandra~Sasha Luccioni, Fran{\c{c}}ois Yvon, Matthias Gall{\'{e}}, Jonathan Tow, Alexander~M. Rush, Stella Biderman, Albert Webson, Pawan~Sasanka Ammanamanchi, Thomas Wang, Beno{\^{\i}}t Sagot, Niklas Muennighoff, Albert~Villanova del Moral, Olatunji Ruwase, Rachel Bawden, Stas Bekman, Angelina McMillan{-}Major, Iz~Beltagy, Huu Nguyen, Lucile Saulnier, Samson Tan, Pedro~Ortiz Suarez, Victor Sanh, Hugo Lauren{\c{c}}on, Yacine Jernite, Julien Launay, Margaret Mitchell, Colin Raffel, Aaron Gokaslan, Adi Simhi, Aitor Soroa, Alham~Fikri Aji, Amit Alfassy, Anna Rogers, Ariel~Kreisberg Nitzav, Canwen Xu, Chenghao Mou, Chris Emezue, Christopher Klamm, Colin Leong, Daniel van Strien, David~Ifeoluwa Adelani, and et~al. 2022.
\newblock \href {https://doi.org/10.48550/ARXIV.2211.05100} {{BLOOM:} {A} 176b-parameter open-access multilingual language model}.
\newblock \emph{CoRR}, abs/2211.05100.

\bibitem[{Schulman et~al.(2017)Schulman, Wolski, Dhariwal, Radford, and Klimov}]{Schulman2017_ppo}
John Schulman, Filip Wolski, Prafulla Dhariwal, Alec Radford, and Oleg Klimov. 2017.
\newblock \href {http://arxiv.org/abs/1707.06347} {Proximal policy optimization algorithms}.
\newblock \emph{CoRR}, abs/1707.06347.

\bibitem[{Song et~al.(2023)Song, Yu, Li, Yu, Huang, Li, and Wang}]{Song2023_pro}
Feifan Song, Bowen Yu, Minghao Li, Haiyang Yu, Fei Huang, Yongbin Li, and Houfeng Wang. 2023.
\newblock \href {https://doi.org/10.48550/ARXIV.2306.17492} {Preference ranking optimization for human alignment}.
\newblock \emph{CoRR}, abs/2306.17492.

\bibitem[{Srivastava et~al.(2023)Srivastava, Huang, Fan, and Yao}]{Srivastava2023_prompt_enhance_zero_shot}
Saurabh Srivastava, Chengyue Huang, Weiguo Fan, and Ziyu Yao. 2023.
\newblock \href {https://doi.org/10.48550/ARXIV.2310.02107} {Instance needs more care: Rewriting prompts for instances yields better zero-shot performance}.
\newblock \emph{CoRR}, abs/2310.02107.

\bibitem[{Taori et~al.(2023)Taori, Gulrajani, Zhang, Dubois, Li, Guestrin, Liang, and Hashimoto}]{Taori2023_alpaca}
Rohan Taori, Ishaan Gulrajani, Tianyi Zhang, Yann Dubois, Xuechen Li, Carlos Guestrin, Percy Liang, and Tatsunori~B. Hashimoto. 2023.
\newblock Stanford {Alpaca}: An instruction-following {LLaMA} model.
\newblock \url{https://github.com/tatsu-lab/stanford_alpaca}.

\bibitem[{Thurstone(1927)}]{Thurstone1927}
Louis~L Thurstone. 1927.
\newblock Psychophysical analysis.
\newblock \emph{The American journal of psychology}, 38(3):368--389.

\bibitem[{Touvron et~al.(2023{\natexlab{a}})Touvron, Lavril, Izacard, Martinet, Lachaux, Lacroix, Rozi{\`{e}}re, Goyal, Hambro, Azhar, Rodriguez, Joulin, Grave, and Lample}]{Touvron2023_llama}
Hugo Touvron, Thibaut Lavril, Gautier Izacard, Xavier Martinet, Marie{-}Anne Lachaux, Timoth{\'{e}}e Lacroix, Baptiste Rozi{\`{e}}re, Naman Goyal, Eric Hambro, Faisal Azhar, Aur{\'{e}}lien Rodriguez, Armand Joulin, Edouard Grave, and Guillaume Lample. 2023{\natexlab{a}}.
\newblock \href {https://doi.org/10.48550/ARXIV.2302.13971} {{LLaMA}: Open and efficient foundation language models}.
\newblock \emph{CoRR}, abs/2302.13971.

\bibitem[{Touvron et~al.(2023{\natexlab{b}})Touvron, Martin, Stone, Albert, Almahairi, Babaei, Bashlykov, Batra, Bhargava, Bhosale, Bikel, Blecher, Canton{-}Ferrer, Chen, Cucurull, Esiobu, Fernandes, Fu, Fu, Fuller, Gao, Goswami, Goyal, Hartshorn, Hosseini, Hou, Inan, Kardas, Kerkez, Khabsa, Kloumann, Korenev, Koura, Lachaux, Lavril, Lee, Liskovich, Lu, Mao, Martinet, Mihaylov, Mishra, Molybog, Nie, Poulton, Reizenstein, Rungta, Saladi, Schelten, Silva, Smith, Subramanian, Tan, Tang, Taylor, Williams, Kuan, Xu, Yan, Zarov, Zhang, Fan, Kambadur, Narang, Rodriguez, Stojnic, Edunov, and Scialom}]{Touvron2023_llama2}
Hugo Touvron, Louis Martin, Kevin Stone, Peter Albert, Amjad Almahairi, Yasmine Babaei, Nikolay Bashlykov, Soumya Batra, Prajjwal Bhargava, Shruti Bhosale, Dan Bikel, Lukas Blecher, Cristian Canton{-}Ferrer, Moya Chen, Guillem Cucurull, David Esiobu, Jude Fernandes, Jeremy Fu, Wenyin Fu, Brian Fuller, Cynthia Gao, Vedanuj Goswami, Naman Goyal, Anthony Hartshorn, Saghar Hosseini, Rui Hou, Hakan Inan, Marcin Kardas, Viktor Kerkez, Madian Khabsa, Isabel Kloumann, Artem Korenev, Punit~Singh Koura, Marie{-}Anne Lachaux, Thibaut Lavril, Jenya Lee, Diana Liskovich, Yinghai Lu, Yuning Mao, Xavier Martinet, Todor Mihaylov, Pushkar Mishra, Igor Molybog, Yixin Nie, Andrew Poulton, Jeremy Reizenstein, Rashi Rungta, Kalyan Saladi, Alan Schelten, Ruan Silva, Eric~Michael Smith, Ranjan Subramanian, Xiaoqing~Ellen Tan, Binh Tang, Ross Taylor, Adina Williams, Jian~Xiang Kuan, Puxin Xu, Zheng Yan, Iliyan Zarov, Yuchen Zhang, Angela Fan, Melanie Kambadur, Sharan Narang, Aur{\'{e}}lien Rodriguez, Robert Stojnic, Sergey Edunov,
  and Thomas Scialom. 2023{\natexlab{b}}.
\newblock \href {https://doi.org/10.48550/ARXIV.2307.09288} {Llama 2: Open foundation and fine-tuned chat models}.
\newblock \emph{CoRR}, abs/2307.09288.

\bibitem[{Train(2003)}]{train_discrete_2003}
Kenneth Train. 2003.
\newblock \href {https://doi.org/10.1017/CBO9780511753930} {\emph{Discrete Choice Method With Simulation}}.
\newblock Cambridge University Press.

\bibitem[{Xu et~al.(2023)Xu, Kim, Sharaf, and Awadalla}]{Xu2023_alma}
Haoran Xu, Young~Jin Kim, Amr Sharaf, and Hany~Hassan Awadalla. 2023.
\newblock \href {https://doi.org/10.48550/ARXIV.2309.11674} {A paradigm shift in machine translation: Boosting translation performance of large language models}.
\newblock \emph{CoRR}, abs/2309.11674.

\bibitem[{Yuan et~al.(2023)Yuan, Yuan, Tan, Wang, Huang, and Huang}]{Yuan2023_rrhf}
Zheng Yuan, Hongyi Yuan, Chuanqi Tan, Wei Wang, Songfang Huang, and Fei Huang. 2023.
\newblock \href {https://doi.org/10.48550/ARXIV.2304.05302} {{RRHF:} rank responses to align language models with human feedback without tears}.
\newblock \emph{CoRR}, abs/2304.05302.

\bibitem[{Zeng et~al.(2023)Zeng, Meng, Yin, and Zhou}]{Zeng2023_tim}
Jiali Zeng, Fandong Meng, Yongjing Yin, and Jie Zhou. 2023.
\newblock Tim: Teaching large language models to translate with comparison.
\newblock \emph{arXiv preprint arXiv:2307.04408}.

\bibitem[{Zhang et~al.(2023{\natexlab{a}})Zhang, Haddow, and Birch}]{Zhang2023}
Biao Zhang, Barry Haddow, and Alexandra Birch. 2023{\natexlab{a}}.
\newblock \href {https://proceedings.mlr.press/v202/zhang23m.html} {Prompting large language model for machine translation: {A} case study}.
\newblock In \emph{International Conference on Machine Learning, {ICML} 2023, 23-29 July 2023, Honolulu, Hawaii, {USA}}, volume 202 of \emph{Proceedings of Machine Learning Research}, pages 41092--41110. {PMLR}.

\bibitem[{Zhang et~al.(2024)Zhang, Gautam, Wang, Alabi, Shen, Klakow, and Mosbach}]{zhang2024impact}
Miaoran Zhang, Vagrant Gautam, Mingyang Wang, Jesujoba~O. Alabi, Xiaoyu Shen, Dietrich Klakow, and Marius Mosbach. 2024.
\newblock \href {http://arxiv.org/abs/2402.12976} {The impact of demonstrations on multilingual in-context learning: A multidimensional analysis}.

\bibitem[{Zhang et~al.(2023{\natexlab{b}})Zhang, Rajabi, Duh, and Koehn}]{Zhang2023_mt_qlora}
Xuan Zhang, Navid Rajabi, Kevin Duh, and Philipp Koehn. 2023{\natexlab{b}}.
\newblock \href {https://aclanthology.org/2023.wmt-1.43} {Machine translation with large language models: Prompting, few-shot learning, and fine-tuning with {QLoRA}}.
\newblock In \emph{Proceedings of the Eighth Conference on Machine Translation}, pages 466--479, Singapore. Association for Computational Linguistics.

\bibitem[{Zheng et~al.(2023)Zheng, Chiang, Sheng, Zhuang, Wu, Zhuang, Lin, Li, Li, Xing, Zhang, Gonzalez, and Stoica}]{Zheng2023_vicuna}
Lianmin Zheng, Wei{-}Lin Chiang, Ying Sheng, Siyuan Zhuang, Zhanghao Wu, Yonghao Zhuang, Zi~Lin, Zhuohan Li, Dacheng Li, Eric~P. Xing, Hao Zhang, Joseph~E. Gonzalez, and Ion Stoica. 2023.
\newblock \href {https://doi.org/10.48550/ARXIV.2306.05685} {Judging {LLM}-as-a-judge with {MT-Bench} and {Chatbot Arena}}.
\newblock \emph{CoRR}, abs/2306.05685.

\bibitem[{Zhou et~al.(2023)Zhou, Liu, Xu, Iyer, Sun, Mao, Ma, Efrat, Yu, Yu et~al.}]{Zhou2023_lima}
Chunting Zhou, Pengfei Liu, Puxin Xu, Srini Iyer, Jiao Sun, Yuning Mao, Xuezhe Ma, Avia Efrat, Ping Yu, Lili Yu, et~al. 2023.
\newblock Lima: Less is more for alignment.
\newblock \emph{arXiv preprint arXiv:2305.11206}.

\bibitem[{Zhu et~al.(2023)Zhu, Shen, Mosbach, Stephan, and Klakow}]{Zhu2023_weaker_than_you_think}
Dawei Zhu, Xiaoyu Shen, Marius Mosbach, Andreas Stephan, and Dietrich Klakow. 2023.
\newblock \href {https://doi.org/10.18653/v1/2023.acl-long.796} {Weaker than you think: A critical look at weakly supervised learning}.
\newblock In \emph{Proceedings of the 61st Annual Meeting of the Association for Computational Linguistics (Volume 1: Long Papers)}, pages 14229--14253, Toronto, Canada. Association for Computational Linguistics.

\end{thebibliography}

\clearpage
\appendix

\section{Incorporating multiple preferences with distance information}
\label{sec:appx:incoerpate_with_more_than_2_preferences}
In Section~\ref{sec:preference_learning}, we demonstrated how the distance information of two preferences can be integrated into preference modeling, as illustrated in Equation~\ref{eq:preference-distance-binary-case}. A similar analysis can be done for the Plackett-Luce ranking model to incorporate distance metrics across multiple preferences. Specifically, we model the probability of a particular ordering $X_1, \cdots, X_L$ as follows:
\begin{align}
P(X_1 \geq X_2 \cdots & \geq X_L) \nonumber \\
& = \prod_{i=1}^{L-1} P_i(X_i > X_j, \forall j>i) \nonumber 
\end{align}
For each distribution $P_i$,  let $X_j = s_j + \varepsilon_j$ for $j\geq i$,  with $\varepsilon_j$ $\sim$ standard Gumbel and independent so that (following~\citet{train_discrete_2003}, Section 3)
\begin{equation*}
P_i(X_i > X_j, \forall j>i) = \frac{e^{s_i}}{ \sum_{j\geq i} e^{s_j}}
\end{equation*}

This ranking can be interpreted as a sequence of $L-1$ independent choices: choose the first item, then choose the second among the remaining alternatives, etc.~\cite{maystre_fast_2015}.    It is usually assumed that each independent choice is made by the same judge whose underlying preferences do not change. If we assume  $s_j = \log \pi_\theta(x, y^j)$ for this judge then Equation~\ref{eq:plackett-luce-rewrite} results.    

Suppose instead that, rather than a single judge, a succession of  $L-1$ different  judges each  make one of the sequence of independent choices.    
The distributions $P_i$ should change to reflect the changing preferences of the judges. In particular, if we introduce the preference distances $d_i^j$ for the $i^{th}$ judge, then we obtain Equation~\ref{eq:plackett-luce-loss-with-distance} if for each $P_i$  the location parameters are  set to $s_j = d_i^j \log \pi_\theta(x, y^j)$ for $j\geq i$.    
We find that this modified version of the Placket-Luce model can work well in practice although we note that these modifications may violate Luce's Choice Axiom~\cite{luce_individual_choice_behaviour_1959,hamilton_many_2023}. 

Consider the case of $L=3$.    The Choice Axiom requires the odds of choosing $X_2$ over $X_3$ are independent of the presence of $X_1$ as an option, i.e. that the odds should not depend on whether this is a choice for the first or the second position  
\begin{equation*}
\frac{P_1(X_2 > X_j, j=1,3)}{P_1(X_3 >  X_j, j=1,2)}  =  \frac{P_2(X_2 > X_3)}{P_2(X_3 > X_2)} 
\end{equation*} 
With the location parameters from above, the Choice Axiom requires
\begin{equation*}
\frac{\pi_\theta(x,y^2)^{d_1^2}}{\pi_\theta(x,y^3)^{d_1^3}} = 
\frac{\pi_\theta(x,y^2)^{d_2^2}}{\pi_\theta(x,y^3)^{d_2^3}} 
\end{equation*}
or that $\pi_\theta(x,y^2)^{(d_1^2 - d_2^2)} = \pi_\theta(x,y^3)^{(d_1^3 - d_2^3)}$.   This holds for the default setting, $d_i^j = 1$, leading to Equation~\ref{eq:plackett-luce-rewrite}, but appears not to hold in general.

We find that the ground truth preference values can be introduced as preference distances in the binary comparison case, but that doing so in the more general case, while useful, may not satisfy the Axiom of Choice.

\section{More details on \mtdata}
\label{sec:appx:more_details_on_our_data}
\subsection{Data Construction}
\label{sec:appx:maple_data_construction}
The source sentences in the training data of \mtdata are sampled from the test sets of WMT20 and WMT21. As mentioned in Section~\ref{sec:preference_collection}, four of the five translations are produced by VicunaMT. Considering that VicunaMT is already a strong MT system, often providing accurate translations free of mistakes, randomly selecting source sentences from WMT data could predominantly yield translations that are trivial for VicunaMT to translate, resulting in the collection of many uninformative samples with high human preference scores. To mitigate this, we prioritize source sentences that present difficulties for VicunaMT. Specifically, we use reference translations as a proxy to assess the quality of the model translations through COMET scores. We give priority to samples where the beam search output falls within a COMET score range of [75,85] and where there is a significant standard deviation in COMET scores among the four translations. Following these criteria, we select 1.1K samples for each translation direction. For the development set in \mtdata, we use monolingual data from News Crawl 2022. The sampling and selection process are the same as that of the training set, except that we do not have reference translations, instead, we use a strong commercial MT system to generate pseudo ``reference'' translations.

\subsection{Scoring Rubric}
\label{sec:appx:scoring_rubric}
The annotators are asked to judge the translation on a scale of 1 to 6, following the guidelines outlined in the following scoring rubric. They can assign scores in increments of 0.2, allowing for more detailed assessments, such as 1.2, 1.4, and so on.
\begin{itemize}
\item \textbf{Score it a 1} when the translation has nothing to do with the source; or when the translation has many unknown words; or when the translation looks like word salad.
\item \textbf{Score it a 2} when you can understand why some of the words in the translation are there, but when the meaning of the source sentence is lost.
\item \textbf{Score it a 3} when you understand why all or almost all the words in the translation are there and when some of the meaning of the source sentence are adequately transferred into the target language, but when the main meaning of the source sentence is lost.
\item \textbf{Score it a 4} when the meaning of the source sentence is generally preserved, but when the translation is mechanical and possibly has vocabulary, grammatical, or date / numbering errors.
\item \textbf{Score it a 5} when the meaning of the source sentence is fully preserved and the translation has no grammatical errors, but when the translation does not sound like the translation a native target language speaker would produce given the style and register of the source sentence.
\item \textbf{Score it a 6} when the translation is perfect in every sense of the word – something a professional translator/interpreter would come up with when she understands well the context in which the source sentence was produced.
\end{itemize}

\begin{figure*}[h]
    \centering
    \includegraphics[width=2\columnwidth]{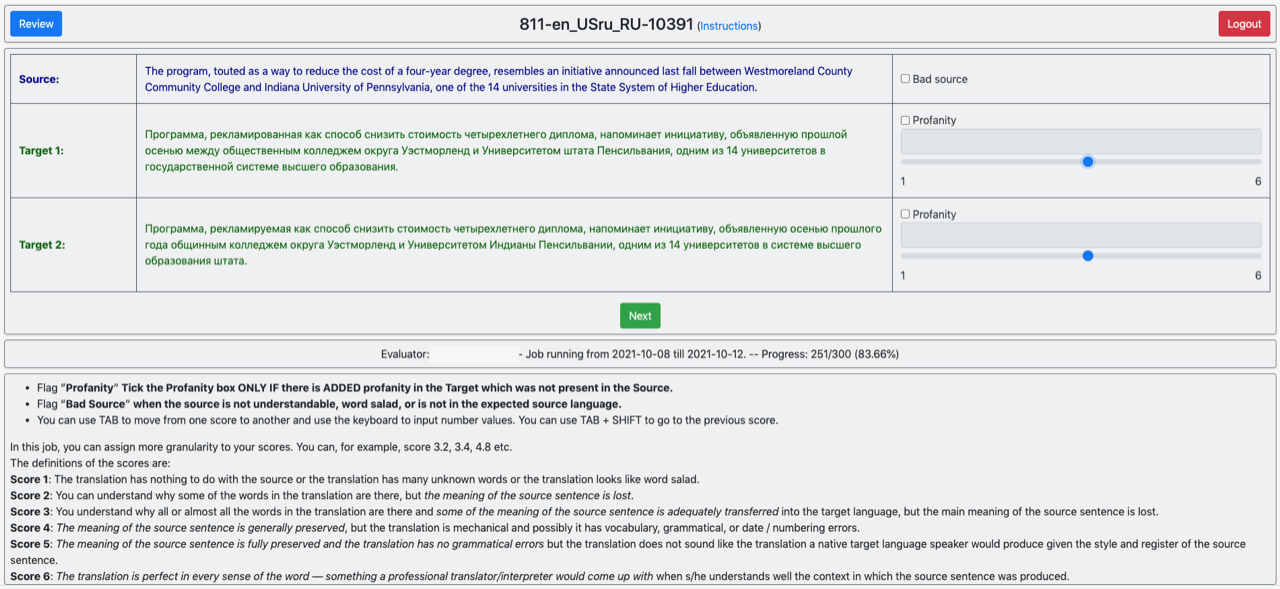}
    \caption{User interface of translation assessment.}
    \label{fig:annotation_ui}
\end{figure*}

\begin{table}[h]
    \centering
    \small
    \begin{tabular}{p{1.5cm}p{4.5cm}}
        \toprule
        Source & Other MPs criticised Twitter for allowing the tweets to \hlblue{remain visible}. \\
        \midrule
        Reference translation & \chinese{其他议员\hlred{也}批评} Twitter\chinese{\hlred{未能及时删贴}。} \\
        & \textit{(Other MPs have \hlred{also} criticized Twitter for \hlred{failing to promptly delete tweets in time}.) }\\
        \midrule
        Best of five translation & \chinese{其他议员批评了推特允许这些推文\hlgreen{仍然可见}。} \\
        & \textit{(Other MPs have criticized Twitter for allowing these tweets to \hlgreen{remain visible}.)} \\
        \midrule
        \midrule
        Source & \hlblue{When he refused}, the officials tipped his cart over, destroying all the eggs, the boy alleged. \\
        \midrule
        Reference translation & \chinese{男孩说，他\hlred{拒绝交出100卢比}后，那些官员就把他的小车掀翻，把所有鸡蛋砸碎。} \\
        & \textit{(The boy said that after he \hlred{refused to hand over 100 rupees}, the officials overturned his car and smashed all the eggs.) }\\
        \midrule
        Best of five translation & \chinese{\hlgreen{当他拒绝时}，官员将他的车子推倒，破坏了所有的蛋，男孩称。} \\
        & \textit{(\hlgreen{When he refused}, officials pushed his car over and broke all the eggs, the boy said.)} \\
        \bottomrule
    \end{tabular}
    \caption{Two additional examples showing the reference translations can be less accurate than the best model prediction.}
    \label{tab:model-better-than-reference2}
\end{table}

\subsection{Annotation UI}
The UI shows the different translations in a blind and randomized order. All translations are scored simultaneously. A screenshot of the UI is shown in Figure~\ref{fig:annotation_ui}.

\subsection{More Examples}
\label{sec:appx:more_details_on_our_data_examples}
Table~\ref{tab:model-better-than-reference2} shows two additional examples in which the model's translation scores higher than the reference translation. This once again highlights the presence of noise in parallel datasets.

\begin{table*}[]
\centering
\begin{tabular}{@{}lcccccc@{}}
\toprule
\multicolumn{1}{@{}c}{} & \multicolumn{1}{@{}c}{Training stage} & \multicolumn{1}{@{}c}{Data source} & \multicolumn{4}{@{}c}{Number of samples} \\
                            & \multicolumn{1}{l}{}       & \multicolumn{1}{l}{} & \multicolumn{1}{l}{\texttt{de}$\rightarrow$\texttt{en}} & \multicolumn{1}{l}{\texttt{en}$\rightarrow$\texttt{de}} & \multicolumn{1}{l}{\texttt{en}$\rightarrow$\texttt{zh}} & \multicolumn{1}{l}{\texttt{zh}$\rightarrow$\texttt{en}} \\ \midrule
\multirow{4}{*}{Training}   & \multirow{3}{*}{SFT stage} & WMT17                & 3004                      & 3004                      & 2001                      & 2001                      \\
                            &                            & WMT18                & 2998                      & 2998                      & 3981                      & 3981                      \\
                            &                            & WMT19                & 2000                      & 1997                      & 1997                      & 2000                      \\ \cmidrule(l){2-7} 
                            & PL stage                   & \mtdata                & 1100                      & 1100                      & 1100                      & 1100                      \\ \midrule
\multirow{2}{*}{Validation} & SFT stage                  & WMT21                & 1000                      & 1002                      & 1002                      & 1948                      \\ \cmidrule(l){2-7} 
                            & PL stage                   & WMT20 \& 21$^\ast$          & 500                       & 500                       & 500                       & 500                       \\ \midrule
\multirow{2}{*}{Test}       & \multirow{2}{*}{-}         & WMT22                & 1984                      & 2037                      & 2037                      & 1875                      \\
                            &                            & FLORES-200           & 1012                      & 1012                      & 1012                      & 1012                      \\ \midrule
Preference testing & - & \mtdata-{dev} & 217 & 195 & 208 & 180  \\ 
                            \bottomrule
\end{tabular}
\caption{Datasets used for training, validation and testing. $^\ast$: a subset WMT20 and WMT21 is used.}
\label{tab:all-data-statistics}
\end{table*}

\section{More implementation details}

\subsection{Dataset statistics}
\label{sec:appx:dataset_statistics}

The data statistics are presented in Table~\ref{tab:all-data-statistics}. We use different validation sets in different training stages because \mtdata contains a subset of the parallel data in WMT20/21.

\begin{table*}[htb!]
    \centering
    \begin{tabular}
    { p{420pt} }
    \toprule
        \textbf{Instruction pool} \\
        \midrule
        \texttt{Translate the following text from \colorbox{orange!30}{[SRC]} to \colorbox{blue!30}{[TGT]}:}
         \\ \hline

         \texttt{Please provide the \colorbox{blue!30}{[TGT]} translation for the following text}
         \\ \hline

        \texttt{Convert the subsequent sentences from \colorbox{orange!30}{[SRC]} into \colorbox{blue!30}{[TGT]}:}
         \\ \hline
        
        \texttt{Render the listed sentences in \colorbox{blue!30}{[TGT]} from their original \colorbox{orange!30}{[SRC]} form:}
         \\ \hline

         \texttt{Transform the upcoming sentences from \colorbox{orange!30}{[SRC]} language to \colorbox{blue!30}{[TGT]} language:}
         \\ \hline

         \texttt{Translate the given text from \colorbox{orange!30}{[SRC]} to \colorbox{blue!30}{[TGT]}:}
         \\ \hline
         \texttt{Turn the following sentences from their \colorbox{orange!30}{[SRC]} version to the \colorbox{blue!30}{[TGT]} version:}
         \\ \hline
         \texttt{Adapt the upcoming text from \colorbox{orange!30}{[SRC]} to \colorbox{blue!30}{[TGT]}:}
         \\ \hline
         \texttt{Transpose the next sentences from the \colorbox{orange!30}{[SRC]} format to the \colorbox{blue!30}{[TGT]} format.}
         \\ \hline
         \texttt{Reinterpret the ensuing text from \colorbox{orange!30}{[SRC]} to \colorbox{blue!30}{[TGT]} language.}
         \\ \hline
         \texttt{Modify the forthcoming sentences, converting them from \colorbox{orange!30}{[SRC]} to \colorbox{blue!30}{[TGT]}.}
         \\ \hline
         \texttt{What is the meaning of these sentences when translated to \colorbox{blue!30}{[TGT]}?}
         \\ \hline
         \texttt{In the context of \colorbox{blue!30}{[TGT]}, what do the upcoming text signify? The text is:}
         \\ \hline
         \texttt{How would you express the meaning of the following sentences in \colorbox{blue!30}{[TGT]}?}
         \\ \hline
         \texttt{What is the significance of the mentioned sentences in \colorbox{blue!30}{[TGT]}?}
         \\ \hline
         \texttt{In \colorbox{blue!30}{[TGT]}, what do the following text convey?}
         \\ \hline
         \texttt{When translated to \colorbox{blue!30}{[TGT]}, what message do these sentences carry?}
         \\ \hline
         \texttt{What is the intended meaning of the ensuing sentences in \colorbox{blue!30}{[TGT]}?}
         \\ \hline
         \texttt{How should the following sentences be comprehended in \colorbox{blue!30}{[TGT]}?}
         \\ \hline
         \texttt{In terms of \colorbox{blue!30}{[TGT]}, what do the next sentences imply?}
         \\ \hline
         \texttt{Kindly furnish the \colorbox{blue!30}{[TGT]} translation of the subsequent sentences.}
         \\ \hline
         \texttt{Could you supply the \colorbox{blue!30}{[TGT]} translation for the upcoming sentences?}
         \\ \hline
         \texttt{Please offer the \colorbox{blue!30}{[TGT]} rendition for the following statements.}
         \\ \hline
         \texttt{I'd appreciate it if you could present the \colorbox{blue!30}{[TGT]} translation for the following text:}
         \\ \hline
         \texttt{Can you deliver the \colorbox{blue!30}{[TGT]} translation for the mentioned sentences?}
         \\ \hline
         \texttt{Please share the \colorbox{blue!30}{[TGT]} version of the given sentences.}
         \\ \hline
         \texttt{It would be helpful if you could provide the \colorbox{blue!30}{[TGT]} translation of the ensuing sentences.}
         \\ \hline
         \texttt{Kindly submit the \colorbox{blue!30}{[TGT]} interpretation for the next sentences.}
         \\ \hline
         \texttt{Please make available the \colorbox{blue!30}{[TGT]} translation for the listed sentences.}
         \\ \hline
         \texttt{Can you reveal the \colorbox{blue!30}{[TGT]} translation of the forthcoming sentences?}
         \\ \hline
         \texttt{Translate from \colorbox{orange!30}{[SRC]} to \colorbox{blue!30}{[TGT]}:}
         \\
    \bottomrule
    \end{tabular}
    \caption{An instruction pool containing 31 MT prompts. An instruction is randomly sampled from this pool to form a training sample. At inference time, the first instruction is always used. \colorbox{orange!30}{[SRC]} and \colorbox{blue!30}{[TGT]} will be replaced by the source and target language, respectively.}
    \label{tab:instruction_pool}
\end{table*}

\begin{table*}[htb!]
    \centering
    \begin{subtable}{\textwidth}
        \centering
        \begin{tabular}{lc}
            \toprule
            \textbf{Model} & \textbf{Instruction template} \\
            \midrule
            Vicuna & \texttt{\colorbox{red!30}{USER: }\colorbox{blue!30}{[MT Instruction]}\colorbox{red!30}{\textbackslash nASSISTANT:\textbackslash n}} \\ \hline
            Mistral-Instruct & \texttt{\colorbox{red!30}{[INST] }\colorbox{blue!30}{[MT Instruction]}\colorbox{red!30}{\textbackslash n[\textbackslash INST]}}\\ \hline
            BLOOMZ & \texttt{\colorbox{red!30}{USER: }\colorbox{blue!30}{[MT Instruction]}\colorbox{red!30}{\textbackslash nASSISTANT:\textbackslash n}} \\
            \bottomrule
        \end{tabular}
        \caption{}
    \end{subtable}

    \bigskip

    \begin{subtable}{\textwidth}
        \centering
        \begin{tabular}{l}
            \toprule
            \textbf{Example} \\
            \midrule
            \texttt{\colorbox{red!30}{USER: }\colorbox{blue!30}{Translate the following text from English to German: }\colorbox{green!30}{Hello, world.}} \\
            \texttt{\colorbox{red!30}{\textbackslash nASSISTANT:\textbackslash n}\colorbox{green!30}{Hallo, Welt.}} \\
            \bottomrule
        \end{tabular}
        \caption{}
    \end{subtable}

    \caption{(a) Instruction template used for Vicuna, Mistral-Instruct, and BLOOMZ. Raw template is marked in \colorbox{red!30}{red}. BLOOMZ shares the same template as Vicuna at the SFT and PL stage. When performing BLOOMZ on zero-shot tasks, we directly use the first instruction from Table~\ref{tab:instruction_pool} without any instruction template. (b) An example that converts the raw input (marked in \colorbox{green!30}{green}) to the final input.}
    \label{tab:vicuna-mistral-bloomz-prompt-example}
\end{table*}

\subsection{Prompt format}
\label{sec:appx:instruction_pool}
For each source sentence, we attach a MT instruction asking the LLM to generate the translation. The MT instructions come from a instruction pool based on the list of MT instructions released by~\cite{Jiao2023_parrot}\footnote{\url{https://github.com/wxjiao/ParroT}}. We list all 31 instructions in our instruction pool in Table~\ref{tab:instruction_pool}. During training (in both SFT and PL stages), an instruction is randomly sampled from the instruction pool. During evaluation, the first instruction from Table~\ref{tab:instruction_pool} is always used. In addition to instructions, instruction-tuned models like Vicuna requires specific prompt formats. Table~\ref{tab:vicuna-mistral-bloomz-prompt-example} presents a depiction of the conversion process from raw data points to the final model input.

\subsection{Hyper-parameter search }
Hyper-parameter search is done for $\beta \in [0.0, 0.05, 0.1]$, and best values are selected according to the validation loss.

\subsection{Evaluation packages}
We use the \texttt{Unbabel/wmt22-comet-da} model\footnote{\url{https://github.com/Unbabel/COMET}} to compute the COMET scores and sacreBLEU\footnote{\url{https://github.com/mjpost/sacrebleu}} for computing BLEU scores. The signature of the sacreBLEU package is \texttt{nrefs:1, case:mixed, eff:no, tok:13a, smooth:exp, version:2.0.0} for all translation directions but \texttt{en}$\rightarrow$\texttt{zh}, in which we use \texttt{tok:zh}.

\subsection{Hardware specifications and runtime}
All experiments are either run on a host with eight NVIDIA A100-40GB GPUs or with eight H100-80GB GPUs. Mixed precision with bfloat16 is used in both SFT and PL. Deepspeed\footnote{\url{https://github.com/microsoft/DeepSpeed}} zero-stage 3 is used when running PL with five preference samples. Each experiment runs no longer than 15 minutes on H100 GPUs.

\begin{table}[ht!]
\resizebox{\columnwidth}{!}{
\begin{tabular}{@{}lccccc@{}}
\toprule
                         & \multicolumn{1}{c}{\texttt{de}$\rightarrow$\texttt{en}} & \multicolumn{1}{c}{\texttt{en}$\rightarrow$\texttt{de}} & \texttt{en}$\rightarrow$\texttt{zh}                     & \multicolumn{1}{c}{\texttt{zh}$\rightarrow$\texttt{en}} & Avg.  \\ \midrule
                         & \multicolumn{5}{c}{\textit{WMT22}}                                                                                             \\
BLOOM      & 1.51   & 0.53 & 1.74   & 5.43 & 2.30 \\
\multicolumn{1}{@{}r}{\textit{+SFT}} & 23.73 & 16.15 & 35.15  & 21.64  & 24.17 \\
\rowcolor{blue!5}
BLOOMZ   & 21.59 & 6.79 & 28.72    & 18.54   & 18.91 \\
\rowcolor{blue!5}
\multicolumn{1}{@{}r}{\textit{+SFT}} & 23.89  & 16.79 & \textbf{35.41}  & 21.01    & 24.28 \\
Mistral &   4.32 &      2.65    &   4.93    &    7.01       &   4.73    \\
\multicolumn{1}{@{}r}{\textit{+SFT}} & \textbf{29.39}    & 24.60  &  31.51   & \textbf{22.09}   & \textbf{26.90} \\
\rowcolor{blue!5}
Mistral-Ins.   & 28.04 & 21.27 & 21.85 & 17.77  & 22.23   \\
\rowcolor{blue!5}
\multicolumn{1}{@{}r}{\textit{+SFT}} & 28.26   & 24.61  & 31.90 & 20.60   & 26.35 \\
LLaMA-1  &  6.30  &     4.00    & 0.88  &    3.01     &   3.55    \\
\multicolumn{1}{@{}r}{\textit{+SFT}} & 28.28     & 19.09   & 25.31 & 20.27     & 23.24 \\
\rowcolor{blue!5}
Vicuna    & 26.16   &        22.11       & 26.26  &          13.91        &   22.11    \\
\rowcolor{blue!5}
\multicolumn{1}{@{}r}{\textit{+SFT}} &  29.26   & \textbf{25.70}  &  29.98   &  20.61   & 26.39 \\
\midrule
& \multicolumn{5}{c}{\textit{FLORES-200}}     \\ 
BLOOM                    & 3.88          & 1.48      & 7.00      & 3.75       & 4.03           \\
\multicolumn{1}{@{}r}{\textit{+SFT}} & 31.85    & 16.26       & \textbf{34.66}    &   \textbf{23.78}            & 26.64                      \\
Mistral    &      3.58         &        1.37          &   0.16       &   1.06      &    1.54       \\
\multicolumn{1}{@{}r}{\textit{+SFT}} &  40.48  &     29.18          &    29.43     &   24.67     &  30.94                \\
\rowcolor{blue!5}
Mistral-Ins.   &     36.81         &     25.64           &   19.81   &  19.25     &         25.38         \\
\rowcolor{blue!5}
\multicolumn{1}{@{}r}{\textit{+SFT}} &   39.16 &  27.79     &     29.77      &      23.10       &   29.96        \\
LLaMA-1  &  4.08   &     2.80       & 1.73     &   1.60                    &         2.55                                                         \\
\multicolumn{1}{@{}r}{\textit{+SFT}} &    40.70      &      29.95      &      20.21       &    20.66        &              27.88                   \\
\rowcolor{blue!5}
Vicuna                   &  35.07      & 26.86   &     26.09      &        17.53      &    26.39     \\
\rowcolor{blue!5}
\multicolumn{1}{@{}r}{\textit{+SFT}} &    \textbf{41.90}       &  \textbf{30.63}        &      28.52      &        23.34          &               \textbf{31.10}                 \\
\bottomrule
\end{tabular}
}
\caption{Model performance (in BLEU score) before and after performing SFT on parallel data. Rows in blue indicate instruction-tuned LLMs. Best results are in \textbf{bold}. Instruction-tuned LLMs perform well even without SFT. Raw LLMs benefits the most from SFT. We exclude BLOOMZ on FLORES-200 as it is a part of BLOOMZ's training data.}
\label{tab:sft-performance-wmt22-bleu}
\end{table}

\section{SFT results in BLEU score}
\label{sec:appx:more_results_in_sft_stage}
We present model performance after SFT stage measured by BLEU score in Table~\ref{tab:sft-performance-wmt22-bleu}. While the general trend remains consistent in comparison to the performance evaluated by COMET, there are some exceptions. For example, although VicunaMT still achieves the top average score on FLORES-200, it is outperformed by MistralMT (i.e., Mistral + SFT) on WMT22.

\begin{table*}[]
\resizebox{2\columnwidth}{!}{
\begin{tabular}{@{}lcccccccccc@{}}
\toprule
\multicolumn{1}{c}{\multirow{2}{*}{System}} & \multicolumn{5}{c}{WMT22}                                                                                             & \multicolumn{5}{c}{FLORES-200}          \\ \cmidrule(lr){2-6} \cmidrule(lr){7-11}
\multicolumn{1}{c}{}                        & \texttt{de}$\rightarrow$\texttt{en}                   & \texttt{en}$\rightarrow$\texttt{de}                     & \texttt{en}$\rightarrow$\texttt{zh}                     & \texttt{zh}$\rightarrow$\texttt{en}                     & Avg.  & \texttt{de}$\rightarrow$\texttt{en} & \texttt{en}$\rightarrow$\texttt{de} & \texttt{en}$\rightarrow$\texttt{zh} & \texttt{zh}$\rightarrow$\texttt{en}& Avg.  \\
\midrule
\multicolumn{11}{@{}c@{}}{\textit{Commercial \& LLaMA-2-7B based MT systems}} \\
ChatGPT\textsubscript{(3.5-turbo-0613)}  &   33.13      &   33.56        &   44.59        &   25.63       &   31.62    &    43.06   &    40.07     &    45.69   &   25.57    &    36.55     \\
GPT-4\textsubscript{(gpt-4-0613)} &   33.72   &   34.84        &  42.75               &  26.33                  & 34.41   &  43.79    &    41.81    &    46.10     &   27.39         &    39.77     \\
ALMA-7B\textsubscript{(LLaMA-2)}  & 29.49 & 30.31 & 36.48& 23.52 & 29.95 &     -$^\otimes$   &   -$^\otimes$    &  -$^\otimes$     &   -$^\otimes$    &   -$^\otimes$    \\ \midrule
\multicolumn{11}{@{}c@{}}{\textit{BLOOMZ-mt-7B based LLMs}} \\
ParroT\textsubscript{(BLOOMZ-mt)}  & 24.90   &  20.50   & 34.50 & 22.70  & 25.65 & -$^\ast$       & -$^\ast$     & -$^\ast$     & -$^\ast$     & -$^\ast$     \\
TIM\textsubscript{(BLOOMZ-mt)}     &  24.31  &  20.63 &  37.20 & 23.42  & 26.39 & -$^\ast$       & -$^\ast$     & -$^\ast$     & -$^\ast$     & -$^\ast$     \\
SWIE\textsubscript{(BLOOMZ-mt)}   &  25.95   &  21.83    & 36.88   &  23.33 & 27.00 & -$^\ast$       & -$^\ast$     & -$^\ast$     & -$^\ast$     & -$^\ast$     \\
\noalign{\vskip 0.2ex}\hdashline\noalign{\vskip 0.2ex}
\multicolumn{11}{@{}c@{}}{\textit{LLaMA-1-7B based LLMs}} \\
ParroT\textsubscript{(LLaMA-1)}   &  27.30   &  26.10  & 30.30   &  20.20  & 25.98 &  39.40 &  30.70 &  29.10 &  21.30 &  32.38 \\
TIM\textsubscript{(LLaMA-1)}      &  27.91      &  25.02  &  30.07     &  19.33    & 25.58 &   39.15  & 29.31  &  28.43 &  22.30 &  29.80 \\
SWIE\textsubscript{(LLaMA-1)}  &  30.48 &  \textbf{27.10}   &  31.08    &  \textbf{21.19}   & \textbf{27.47}  &  40.20  &  \textbf{31.41}  &  29.07  &  21.59     &  30.57     \\
VicunaMT\textsubscript{(LLaMA-1)}    &  29.26      & 25.70     &  29.98    &  20.61   & 26.39 &  \textbf{41.90}       &  30.63        &      28.52      &        \textbf{23.34}          &               \textbf{31.10}   \\
+ REF  &    \textbf{31.12} &    24.72   &      30.07    &    20.38    &   26.58    &  39.03  &   29.36 & 28.87 &  22.84 & 30.03    \\
+ BEST      & 29.44  &    24.93        &      30.91    &        20.39       &    26.16   &  41.29    &   29.34   &  30.07    &   23.48  &   31.05    \\
+ PL         &    30.63     &      24.63        &     \textbf{31.52}    &    20.44      &    26.81  & 40.07     &   29.33    &   \textbf{30.50}    &   21.99   &  30.47    \\ 
\bottomrule
\end{tabular}
}
\caption{Model performance in BLEU scores. Best results with LLaMA-1 based models are in \textbf{bold}. $^\otimes$: LLaMA-2 based models were not evaluated due to license constraints. WMT22 results are extracted from the original paper. $^\ast$: BLOOMZ-family models use FLORES-200 for training.}
\label{tab:preference_learning_results_bleu}
\end{table*}

\section{Model comparison in BLEU score}
\label{sec:appx:model_comparison_bleu_score}
We present model performance measured by BLEU score in Table~\ref{tab:preference_learning_results_bleu}. In this case, there is no clear winner. Interestingly, VicunaMT+PL attains lower BLEU scores than VicunaMT on \texttt{en}$\rightarrow$\texttt{de} and \texttt{zh}$\rightarrow$\texttt{en} when evaluated on WMT22. However, both COMET score and our human evaluation in Table~\ref{tab:humaneval} show the opposite, highlighting that BLEU scores may less correlated to human judgement, as also noticed in~\cite{Freitag2022}.

\begin{table}[ht!]
\resizebox{\columnwidth}{!}{
\begin{tabular}{@{}lccccc@{}}
\toprule
                         & \multicolumn{5}{c}{\textit{WMT22}} \\
                         & \multicolumn{1}{c}{\texttt{de}$\rightarrow$\texttt{en}} & \multicolumn{1}{c}{\texttt{en}$\rightarrow$\texttt{de}} & \texttt{en}$\rightarrow$\texttt{zh}                     & \multicolumn{1}{c}{\texttt{zh}$\rightarrow$\texttt{en}} & Avg.  \\ \midrule
                         
BLOOMZ$^\dagger$ & 23.89  & 16.79 & 35.41    & 21.01       & 24.28 \\
+REF & 24.51  & 15.26 & 33.43  & 21.80   & 23.75 \\
+BEST & 23.80  & 16.33 & 34.99  & 21.49     & 24.15 \\
+PL &  \textbf{24.84}  & \textbf{16.81} & \textbf{36.48 }  &\textbf{ 23.15} & \textbf{25.32}  \\
\midrule
Mistral-Ins.$^\dagger$ & 28.26   & 24.61  & 31.90 & 20.60   & 26.35 \\
+REF & \textbf{30.94 }   & \textbf{25.62}   & 31.66 & 21.52   & 27.44 \\
+BEST & 29.76  & 24.30  & 31.12 & 20.83     & 26.50 \\
+PL & 29.32   & 24.78    & \textbf{33.00} & \textbf{21.76 }     & \textbf{27.47} \\
\bottomrule
\end{tabular}
}
\caption{Model performance on WMT22 in BLEU scores. Best results are in \textbf{bold}. $^\dagger$: SFT stage has already been applied to these models.}
\label{tab:preference-data-help-other-LLMs-bleu}
\end{table}

\begin{table}[h]
\resizebox{\columnwidth}{!}{
\begin{tabular}{@{}lccccc@{}}
\toprule
                         & \multicolumn{5}{c}{\textit{FLORES-200}} \\
                         & \multicolumn{1}{c}{\texttt{de}$\rightarrow$\texttt{en}} & \multicolumn{1}{c}{\texttt{en}$\rightarrow$\texttt{de}} & \texttt{en}$\rightarrow$\texttt{zh}                     & \multicolumn{1}{c}{\texttt{zh}$\rightarrow$\texttt{en}} & Avg.  \\ \midrule
& \multicolumn{5}{c}{\textit{COMET}} \\ 
Mistral-Ins.$^\dagger$ & 88.21  &    83.73     &    82.41     &   84.77    &  84.78 \\
+REF & 88.10   & 85.04   & 83.59 & 84.74   & 85.37 \\
+BEST & 88.41  & 84.55  & 83.46 & 84.94     & 85.34 \\
+PL & \textbf{88.56}   & \textbf{84.98}    & \textbf{83.86} & \textbf{85.34}     & \textbf{85.67} \\ \midrule
& \multicolumn{5}{c}{\textit{BLEU}} \\
Mistral-Ins.$^\dagger$ & 39.16 &  27.79     &     29.77      &      23.10       &   29.96        \\
+REF & 38.10   & \textbf{28.39}   & \textbf{31.24} & 23.09   & 30.21 \\
+BEST & 39.35  & 28.33  & 30.46 & 22.98     & 30.28 \\
+PL & \textbf{39.80}   & 27.97    & 31.00 & \textbf{23.44}     & \textbf{30.55} \\
\bottomrule
\end{tabular}
}
\caption{Model performance on FLORES-200 in COMET and BLEU scores. Best results are in \textbf{bold}. $^\dagger$: SFT stage has already been applied to these models.}
\label{tab:preference-data-help-other-LLMs-flores200-comet-bleu}
\end{table}

\section{Data reuse in BLEU score and Results on FLORES-200}
\label{sec:appx:data_reuse_bleu_score}
We reuse \mtdata to enhance BLOOMZMT and MistralInstructMT (i.e., BLOOMZ and MistralInstruct after the SFT stage) and report model performance on WMT22 in BLEU score in Table~\ref{tab:preference-data-help-other-LLMs-bleu}. In addition, we evaluate MistralInstructMT on FLORES and present the results in Table~\ref{tab:preference-data-help-other-LLMs-flores200-comet-bleu}.

\end{document}